\documentclass[12pt]{article}
\usepackage[utf8]{inputenc}
\usepackage[margin = 1in]{geometry}
\usepackage{setspace}
\usepackage{bm}
\usepackage{blindtext}
\usepackage{amsthm, amssymb, amsmath}
\usepackage{mathrsfs}
\usepackage[dvipsnames]{xcolor}
\usepackage{natbib}
\usepackage{enumerate}
\usepackage{enumitem}
\usepackage{bm}
\usepackage{graphicx}
\usepackage{algorithmicx}
\usepackage{algpseudocode}
\usepackage{algorithm}
\usepackage{nicefrac}
\usepackage{pdfpages}

\definecolor{alertcolor}{RGB}{92, 16, 127}
\definecolor{tonycolor}{RGB}{232, 55, 70}

\newcommand{\bdelta}{\bm \delta}
\newcommand{\blambda}{\bm \lambda}
\newcommand{\bY}{\bm Y}
\newcommand{\bX}{\bm X}
\newcommand{\bartMachine}{\texttt{bartMachine}}
\newcommand{\Data}{\mathcal D}

\newcommand{\Bart}{\texttt{BART}}
\newcommand{\Bernoulli}{\operatorname{Bernoulli}}
\newcommand{\Beta}{\operatorname{Beta}}
\newcommand{\Binomial}{\operatorname{Binomial}}
\newcommand{\Birth}{\textnormal{\texttt{BIRTH}}}
\newcommand{\blackboost}{\texttt{blackboost}}
\newcommand{\Branches}{\mathcal B}
\newcommand{\Categorical}{\operatorname{Categorical}}
\newcommand{\Change}{\textnormal{\texttt{CHANGE}}}
\newcommand{\Death}{\textnormal{\texttt{DEATH}}}

\newcommand{\Dirichlet}{\operatorname{Dirichlet}}
\newcommand{\E}{\mathbb E}
\newcommand{\Ell}{\mathscr L}

\newcommand{\Fisher}{\mathcal I}
\newcommand{\Gam}{\operatorname{Gam}}
\newcommand{\iid}{\stackrel{\textnormal{iid}}{\sim}}
\newcommand{\Jags}{\texttt{JAGS}}
\newcommand{\Leaves}{\mathcal L}
\newcommand{\Logistic}{\operatorname{Logistic}}
\newcommand{\mboost}{\texttt{mboost}}
\newcommand{\MSE}{\operatorname{MSE}}

\newcommand{\Nodes}{\mathcal N}
\newcommand{\NOG}{\operatorname{NOG}}
\newcommand{\Normal}{\operatorname{Normal}}
\newcommand{\OFisher}{\mathcal J}
\newcommand{\Poisson}{\operatorname{Poisson}}
\newcommand{\Polya}{Pólya}
\newcommand{\R}{\texttt{R}}
\newcommand{\rbart}{\texttt{rbart}}
\newcommand{\Reals}{\mathbb R}
\newcommand{\RMSE}{\operatorname{RMSE}}
\newcommand{\sH}{\mathcal H}
\newcommand{\sM}{\mathcal M}
\newcommand{\sF}{\mathscr F}
\newcommand{\sigmoid}{\mathfrak s}
\newcommand{\Stan}{\texttt{Stan}}
\newcommand{\Tensorflow}{\texttt{TensorFlow}}
\newcommand{\Tree}{\mathcal T}
\newcommand{\Uniform}{\operatorname{Uniform}}
\newcommand{\Var}{\operatorname{Var}}
\newcommand{\xgboost}{\texttt{xgboost}}

\newtheorem{proposition}{Proposition}

\title{Generalized Bayesian Additive Regression Trees Models: Beyond Conditional Conjugacy}

\author{Antonio R. Linero\thanks{Department of Statistics and Data Sciences, University of Texas at Austin, email: \href{mailto:antonio.linero@austin.utexas.edu}{antonio.linero@austin.utexas.edu}}}

\date{}

\usepackage[colorlinks, citecolor = NavyBlue]{hyperref}

\begin{document}

\maketitle

\begin{abstract}
    Bayesian additive regression trees have seen increased interest in recent years due to their ability to combine machine learning techniques with principled uncertainty quantification. The Bayesian backfitting algorithm used to fit BART models, however, limits their application to a small class of models for which conditional conjugacy exists. In this article, we greatly expand the domain of applicability of BART to arbitrary \emph{generalized BART} models by introducing a very simple, tuning-parameter-free, reversible jump Markov chain Monte Carlo algorithm. Our algorithm requires only that the user be able to compute the likelihood and (optionally) its gradient and Fisher information. The potential applications are very broad; we consider examples in survival analysis, structured heteroskedastic regression, and gamma shape regression. 
\end{abstract}

\doublespacing

\section{Introduction}

Since the introduction of boosting \citep{freund1999short}, algorithms that ensemble shallow decision trees have become a fundamental part of the data science toolkit. A Bayesian framework for ensembling shallow decision trees is the Bayesian additive regression trees (BART) framework of \citet{chipman2010bart}. Some advantages of BART over other machine learning algorithms are that it provides direct uncertainty quantification and can naturally be incorporated into hierarchical models; while there are currently no theoretical guarantees regarding uncertainty quantification, it has been observed that BART performs surprisingly well in practice relative to other attempts at combining machine learning with statistical inference \citep{dorie2019automated}.

A drawback of BART is that one usually needs to tailor it to the problem at hand. Since the initial work of \citet{chipman2010bart}, which developed methods for semiparametric regression and classification, there have been substantial efforts to extend BART to other settings; a limited set of examples include survival analysis \citep{sparapani2016nonparametric, linero2021bayesian}, Poisson regression \citep{murray2020log}, and gamma regression \citep{linero2018shared}. These developments have required either (i) the model to be such that software for normal or probit models can be adapted or (ii) the involvement of experts in BART methodology. 

The difficulty of implementing new BART models stands in stark contrast with the difficulty of implementing new decision tree boosting algorithms, which can be done with very minimal expertise. In particular, given outcomes $\bY = (Y_1, \ldots, Y_N)$, covariate vectors $\bX = (X_1, \ldots, X_N)$, and any utility function $R(\bY \mid \bX, r, \eta) = \sum_{i=1}^N R_\eta\big(Y_i \mid r(X_i)\big)$ with nuisance parameter vector $\eta$, one can construct a \emph{gradient boosting} algorithm \citep{friedman2001greedy} for estimating the function $r(x)$ that only requires users to provide the functions $R_\eta(y \mid \lambda)$, $U_\eta(y \mid \lambda) = \frac{\partial}{\partial \lambda} R_\eta(y \mid \lambda)$, and, optionally, $\OFisher_\eta(y \mid \lambda) = -\frac{\partial}{\partial \lambda} U_\eta(y \mid \lambda)$; for model-based inference with a parametric family $\{f_\eta(\cdot\mid\lambda)\}$, note that we can take $R_\eta(y \mid \lambda) = \log f_\eta(y \mid \lambda)$. Software such as the \R\ packages \xgboost\ and \mboost\ make it straight-forward for users to supply these functions manually, allowing boosting to be applied with arbitrary models and loss functions. This difference between BART and boosting is not because gradient boosting is an intrinsically simpler algorithm, but rather because the complex parts of the algorithm can be abstracted away from the user; in addition to boosting, abstracting away the implementation of complex inference algorithms has been key to the success of applied Bayesian modeling, which has been fueled by packages such as \Stan\ and \Jags.

Our primary aim is to introduce a framework for fitting \emph{generalized BART} models with likelihoods of the form $\prod_{i} f_\eta\{Y_i \mid r(X_i)\}$ that, like gradient boosting, allows us to automate the application of BART to new settings. The main obstacle to this has been the reliance of BART on the generalized Bayesian backfitting algorithm described by \citet{hill2019bayesian}, which requires users to be able specify a prior $\pi_\mu(\mu)$ such that the integrated likelihood $\Lambda = \int \prod_i f_\eta(Y_i \mid \lambda_i + \mu) \, \pi_\mu(\mu) \ d\mu$ can be computed in closed form; this is used to compute a Metropolis-Hastings  acceptance probability for modifying the structure of a tree. Rather than starting from the assumption that  $\Lambda$ is analytically tractable, we instead assume (like boosting) that  $\log f_\eta(y \mid \lambda)$ and its derivatives have been provided; strictly speaking even the derivatives need not be provided, as our algorithm can also be applied by approximating the derivatives with finite differences. Using only this assumption, we construct a generic reversible jump Markov chain Monte Carlo (RJMCMC, \citealp{green1995reversible}) algorithm to sample new tree structures. The jump between dimensions is constructed using a Laplace approximation to ensure that the proposal has a high probability of being accepted. Importantly, our proposal is agnostic to the choice of model and completely free of tuning parameters.

We implement several models to illustrate both the accuracy and flexibility of our approach. We benchmark our algorithm on both semiparametric regression and classification problems, which are handled by existing algorithms; as a bonus, the RJMCMC algorithm avoids any data augmentation \citep{AlbertChib:1993}. We then move on to previously intractable models such as structured variance modeling, accelerated failure time modeling with the log-logistic and generalized gamma distributions, and modeling of the shape parameter in gamma regression. In all cases we find that our RJMCMC algorithm works well.

In Section~\ref{sec:bart} we review the BART models which can currently be fit using existing Bayesian backfitting algorithms. In Section~\ref{sec:rjmcmc} we develop our RJMCMC algorithm for arbitrary generalized BART models. In Section~\ref{sec:applications} we illustrate our approach on a variety of both real and simulated problems. We close in Section~\ref{sec:discussion} with a discussion.

\section{Bayesian Additive Regression Trees}
\label{sec:bart}

\subsection{A Brief Review of BART}
\label{sec:bart-review}

\begin{figure}[t]
    \centering
    \includegraphics[width=0.8\textwidth]{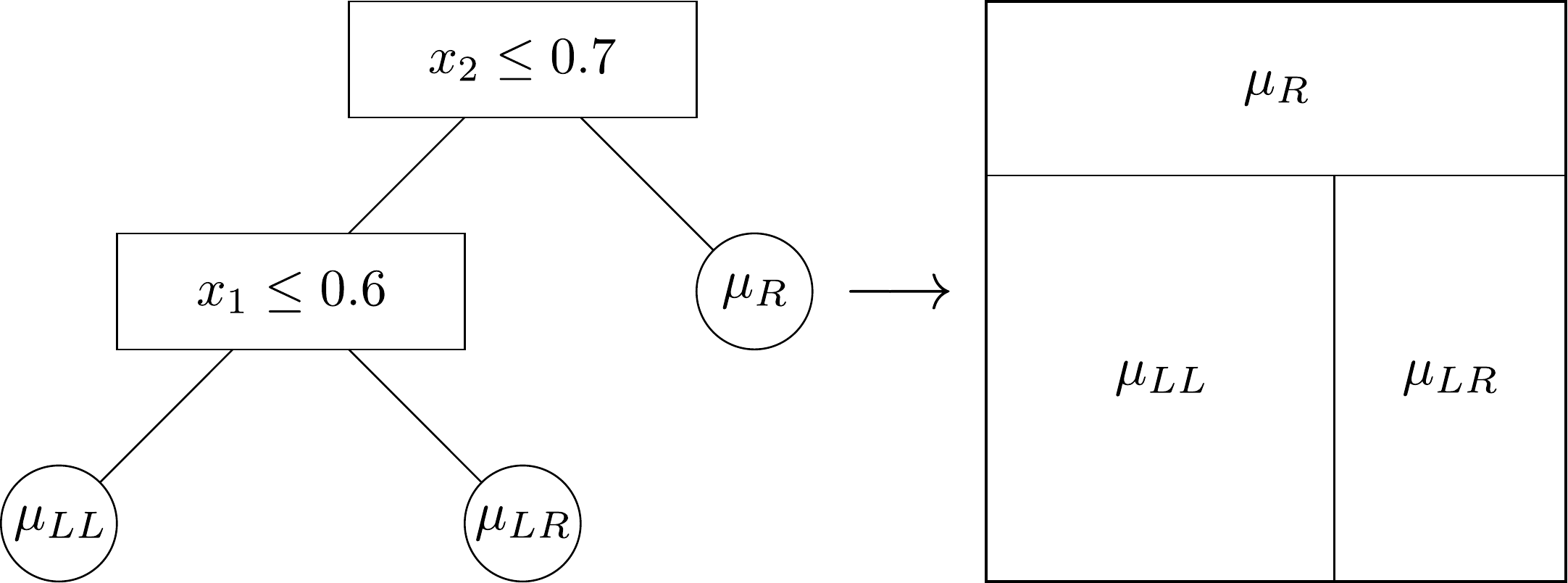}
    \caption{Schematic showing how a regression tree (left) gives rise to a step function of the predictors (right).}
    \label{fig:treefig}
\end{figure}

Suppose we have outcome data $\bY = (Y_1, \ldots, Y_N)$ and covariates $\bX = (X_1, \ldots, X_N)$ where, for simplicity, we assume that $X_i$ takes values in $[0,1]^P$. The Bayesian additive regression trees (BART) model as originally proposed by \citet{chipman2010bart} is a semiparametric regression model of the form
\begin{align}
    \label{eq:semipar-bart}
    Y_i \sim \Normal\{r(X_i), \sigma^2\}
    \qquad\text{where}\qquad
    r(x) = \sum_{t=1}^T g(x ; \Tree_t, \sM_t) \quad (i = 1,\ldots,N),
\end{align}
where $N$ is the sample size. The functions $g(x; \Tree_t, \sM_t)$ are \emph{regression trees} parameterized by a \emph{decision tree} $\Tree_t$ and a collection of predictions for the leaf nodes $\sM_t$. Formally, we define a (binary) decision tree $\Tree$ as a collection of nodes $n \in \Nodes(\Tree)$ where $n$ is a finite (potentially empty) string of the symbols $L$ (left) and $R$ (right). We say that $\ell \in \Nodes(\Tree)$ is a \emph{leaf node} of $\Tree$ if both $\ell L \notin \Nodes(\Tree)$ and $\ell R \notin \Nodes(\Tree)$. Any node $b$ which is not a leaf node is called a \emph{branch node}, and we require that both $bL \in \Nodes(\Tree)$ and $bR \in \Nodes(\Tree)$ for every branch $b$. We let $\Leaves(\Tree)$ and $\Branches(\Tree)$ denote the leaf and branch nodes of $\Tree$ respectively. It will also be convenient for us to define $\NOG(\Tree)$ to be the set of \emph{non-grandparent} nodes, i.e., $\NOG(\Tree) = \{b \in \Branches(\Tree) : bR \in  \Leaves(\Tree) \text{ and } bL  \in \Leaves(\Tree)\}$; for example, the only non-grandparent branch in the tree in Figure~\ref{fig:treefig} is the branch $b = L$.

Associated to each $b \in \Branches(\Tree)$ is a \emph{splitting rule} of the form $[x_{j_b} \le C_b]$. If $x$ is associated to $b$ and $x$ satisfies $b$'s splitting rule then we associate $x$ to $bL$; otherwise, we associate $x$ to $bR$. We write $x \stackrel{\Tree}{\leadsto} n$ (or $x \leadsto n$ when $\Tree$ is clear from context) to denote that $x$ is associated to node $n$ of tree $\Tree$. The collection of predictions can then be defined by $\sM_t = \{\mu_{t\ell} : \ell \in \Leaves(\Tree_t)\}$. By design, the leaf nodes partition the predictor space so that $x \leadsto \ell$ for exactly one $\ell \in \Leaves(\Tree_t)$. Given $(\Tree_t, \sM_t)$ the decision tree outputs the prediction $g(x ; \Tree_t, \sM_t) = \mu_{t\ell}$ if-and-only-if $x \stackrel{\Tree_t}{\leadsto} \ell$. A schematic showing how predictions are generated from a regression tree is given in Figure~\ref{fig:treefig}.

The BART model places independent priors on the regression trees $(\Tree_t, \sM_t) \iid \pi_\Tree(\Tree_t) \, \pi_{\sM}(\sM_t \mid \Tree_t)$. We assume independence across the leaf node parameters, i.e., $\pi_{\sM}(\sM_t \mid \Tree_t) = \prod_{\ell \in \Leaves(\Tree_t)} \, \pi_{\mu}(\mu_{t\ell})$. When possible, $\pi_{\mu}$ is chosen so that it is conditionally conjugate; for the model \eqref{eq:semipar-bart} we take $\pi_\mu(\mu) = \Normal(\mu \mid 0, \sigma^2_\mu)$.

The most common choice of prior for $\pi_{\Tree}(\Tree)$ is a \emph{branching process}: starting at depth $d = 0$, each node of depth $d$ is made a branch node with probability $\rho_d = \gamma (1 + d)^{-\beta}$ and is made a leaf otherwise. This process iterates until all nodes at depth $d$ are leaves. After the shape of the tree is generated, \citet{chipman2010bart} propose generating the splitting rules $[x_{j_b} \le C_b]$ for each $b \in \Branches(\Tree)$ by (i) sampling a decision rule $j_b$ from $\{1,\ldots,P\}$ such that $j_b$ can produce a ``valid'' splitting rule and (ii) sampling $C_b \sim \Uniform(X_{ij} : X_i \leadsto b)$ such that the splitting rule is ``valid''; if no such valid $X_{ij}$ exists, we instead convert the node into a branch and remove all of its descendants. For a rule to be valid, \citet{chipman2010bart} require that the rule associate some minimum number of $X_i$'s to each child node (say, 5). A simple alternative, which we use here, is to simply take $j_b = j$ with some probability $s = (s_1, \ldots, s_P)$ (the simplest option being $s_j = 1/P$) and then sample $C_b \sim \Uniform(L_{bj}, U_{bj})$ where $\prod_{k=1}^P [L_{bk}, U_{bk}]$ is the hyperrectangle in $[0,1]^P$ of points $x$ with $x \leadsto b$. 

\begin{figure}[t]
  \centering
  \includegraphics[width=.9\textwidth]{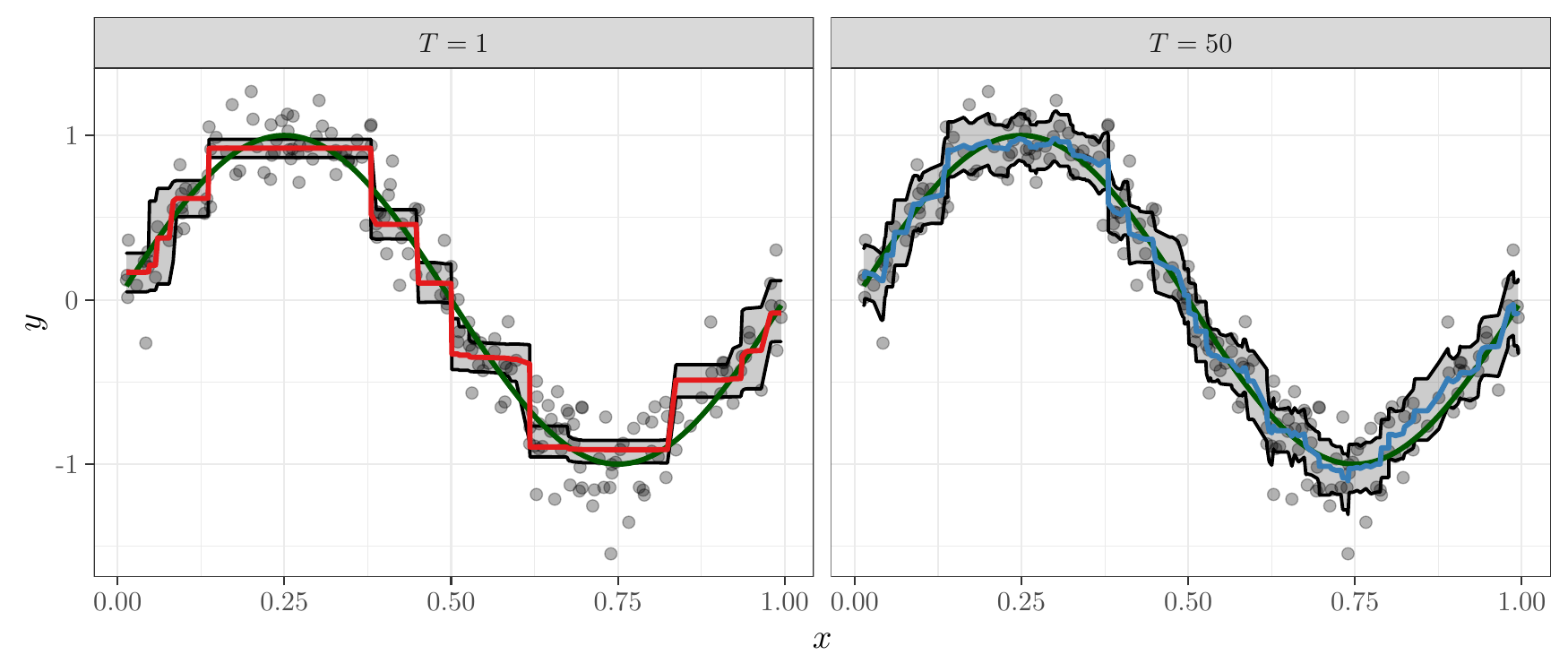}
  \caption{Comparison of the fit of BART with $T = 1$ (left) and $T = 50$ (right) to $Y_i \sim \Normal\{\sin(2\pi X_i), 0.2^2\}$. The function $\sin(2\pi x)$ is given by the dark green line; bands correspond to posterior 95\% credible bands.}
  \label{fig:BartBcartCompare}
\end{figure}

BART improves upon using a single decision tree $r(x) = g(x; \Tree, \sM)$ in several ways. First, as seen in Figure~\ref{fig:BartBcartCompare}, the addition of many decision trees together can \emph{smooth} the estimates of a function; this results in both more accurate predictions and uncertainty quantification. Second, the posterior tends to be easier to explore when many trees are used. Third, as argued heuristically by \citet{chipman2010bart} and rigorously by \citet{rockova2017posterior, linero2017abayesian}, BART models induce a ``shrinkage towards approximately additive models:'' samples of BART from the prior tend to involve, at most, lower-order interactions in the covariates. Outside of highly structured problems (e.g., image or speech recognition), this structure is representative of what one often expects to see in practice; for this reason, BART has been seen to perform very well across many problems in prediction \citep{chipman2010bart}, survival analysis \citep{sparapani2016nonparametric}, and causal inference \citep{hahn2020bayesian, hill2011bayesian}.

\subsection{Generalized BART Models}
\label{sec:generalized-bb}

In this paper we consider BART models in which the function $r(x)$ enters the model in an arbitrary form. Our approach is applicable to any posterior of the form
\begin{align}
  \label{eq:generic}
    \pi(r, \eta \mid \Data)
    \propto
    \exp\left\{\sum_{i=1}^N \log f_\eta\big(Y_i \mid r(X_i)\big)\right\}
    \pi(r) \, \pi(\eta),
\end{align}
where $\log f_\eta(y \mid \lambda)$ is the log-likelihood of some parametric model $\{f_\eta(\cdot \mid \lambda) : \eta \in \sH, \lambda \in \Reals\}$ and $\eta$ is a vector of nuisance parameters. We note, however, that it is straight-forward to replace $\log f_\eta(y \mid \lambda)$ with an arbitrary utility function $R_\eta(y \mid \lambda)$ in our framework.
We say that the model is a \emph{generalized BART model} if $r$ has a BART prior.
We remark that, just as generalized linear models fall outside the ``general linear model,'' generalized BART models are not examples of the \emph{general BART model} described by \citet{tan2019bayesian}. 

The seminal work of \citet{chipman2010bart} develops the semiparametric regression model $f_\sigma(y \mid \lambda) = \Normal(y \mid \lambda, \sigma^2)$ and the Binomial probit regression model $f_n(y \mid \lambda) = \Binomial\big(y \mid n, \Phi(\lambda)\big)$. Several other models have also been developed in this framework, such as the Poisson model $f(y \mid \lambda) = \Poisson(y \mid e^\lambda)$ \citep{murray2020log} and the gamma regression model $f_\alpha(y \mid \lambda) = \Gam(y \mid \alpha, e^\lambda)$ \citep{linero2018shared}. Taking the nuisance parameter $\eta$ to be infinite-dimensional, this also includes several recently proposed BART models for fully-nonparametric regression and survival analysis \citep{henderson2020individualized, george2019fully, li2020adaptive, linero2021bayesian}. 

The need for generic algorithms for fitting generalized BART models is evinced by the fact that, in some cases, the theoretical development of generalized BART has preceded our ability to implement it. For example, \citet{saha2021flexible} proposes and studies BART models in the exponential family $f(y \mid \lambda) = \exp\{\lambda \, T(y) - b(\lambda) + c(y)\}$ without providing algorithms for fitting these models. Instead, prior to this work, implementing new instances of the generalized BART model required researchers to either find clever ways of adapting existing Bayesian backfitting algorithms (e.g., by introducing latent variables as in \citealp{kindo2016bayesian}) or find novel setups for leveraging conjugacy \citep{murray2020log}; both options generally require extensively modifying existing software.


\subsection{Bayesian Backfitting in Generalized BART Models}

Inference in the semiparametric model \eqref{eq:semipar-bart} proceeds by means of a \emph{Bayesian backfitting} algorithm, which iteratively updates the pairs $(\Tree_t, \sM_t)$ for $t = 1,\ldots,T$. To facilitate forthcoming comparisons with our RJMCMC algorithm, we describe the original Bayesian backfitting algorithm of \citet{chipman2010bart} in a slightly unconventional way. In order to update $(\Tree_t, \sM_t)$, we first define $\lambda_i = \sum_{k \ne t} g(X_i ; \Tree_k, \sM_k)$ so that $Y_i \sim \Normal(\lambda_i + \mu_{\ell}, \sigma^2)$ where $\ell$ is the leaf such that $X_i \stackrel{\Tree_t}{\leadsto} \ell$. The full conditional $\pi(\Tree_t \mid \Tree_{-t}, \sM_{-t}, \bX, \bY, \sigma^2)$ of $\Tree_t$ with $\sM_t$ marginalized out is then proportional to
\begin{align}
  \label{eq:bb}
  \begin{split}
    &\pi_\Tree(\Tree_t)
    \prod_{\ell \in \Leaves(\Tree_t)}
    \int \Normal(\mu \mid 0, \sigma^2_\mu)
    \prod_{i: X_i \stackrel{\Tree_t}{\leadsto} \ell} \Normal(Y_i \mid \lambda_i + \mu) \ d\mu.
    \\&\qquad=
    \pi_\Tree(\Tree_t)
    \prod_{\ell \in \Leaves(\Tree_t)}
    \int \Normal(\mu \mid 0, \sigma^2_\mu)
    \prod_{i: X_i \stackrel{\Tree_t}{\leadsto} \ell} \Normal(R_i \mid \mu, \sigma^2) \ d\mu,
  \end{split}
\end{align}
where $R_i$ denotes the \emph{backfit residual} $Y_i - \lambda_i$, $\Tree_{-t} = \{\Tree_k : k \ne t\}$, and $\sM_{-t} = \{\mu_{k\ell} : k \ne t \}$. Importantly, this marginal likelihood can be computed in closed-form due to the conjugacy properties of the normal distribution (see \citealp{kapelner2014bartmachine} for details). This allows us to update $\Tree_t$ using a Metropolis-Hastings algorithm: we sample $\Tree' \sim q(\Tree' \mid \Tree_t)$ from some proposal distribution $q(\cdot\mid\cdot)$ and accept or reject it according to a Metropolis-Hastings ratio based on \eqref{eq:bb}. Generally, the \Birth, \Death, and \Change\ proposals of \citet{chipman1998bayesian} (or the more advanced versions of these moves proposed by \citealp{pratola2016efficient}) are used for $q(\cdot\mid\cdot)$; we discuss variants of these moves in Section~\ref{sec:parameter-expanded}. While BART was initially developed for semiparametric regression, \citet{chipman2010bart} show how to adapt \eqref{eq:semipar-bart} to classification settings using a probit model $Y_i \sim \Bernoulli[\Phi\{r(X_i)\}]$. Inference then proceeds by combining the above Metropolis-Hastings approach with the data augmentation procedure of \citet{AlbertChib:1993}.

While convenient and intuitive, the process of going from $Y_i$ to $R_i$ masks a more general expression which allows the Bayesian backfitting algorithm to be generalized; specifically, for a generic parametric model $f_\eta(y \mid \lambda)$ the relevant conditional distribution is
\begin{align*}
  \pi(\Tree_t \mid \Tree_{-t}, \sM_{-t}, \bY, \bX, \sigma^2)
  \propto
  \pi_\Tree(\Tree_t) \prod_{\ell \in \Leaves(\Tree_t)} \int \pi_\mu(\mu) \prod_{i : X_i \stackrel{\Tree_t}{\leadsto} \ell} f_\eta(Y_i \mid \lambda_i + \mu) \ d\mu.
\end{align*}
We can therefore generalize the Bayesian backfitting algorithm if we can make $\pi_\mu(\mu)$ conjugate to $\prod_i f_\eta(Y_i \mid \lambda_i + \mu)$. A \emph{generalized Bayesian backfitting} algorithm based on this expression is given in Algorithm~\ref{alg:gbb}. Beyond the normal-normal model, this Bayesian backfitting algorithm has been used to implement (i) Poisson loglinear models and multinomial logistic regression \citep{murray2020log}, (ii) gamma regression \citep{linero2018shared}, (iii) nonparametric variance models \citep{pratola2017heteroscedastic}, and (iv) the Cox proportional hazards model \citep{linero2021bayesian}. 
For example, the Poisson loglinear model takes $f(Y_i \mid \lambda_i + \mu) = \Poisson(Y_i \mid e^{\lambda_i + \mu})$, for which the log-gamma distribution $\mu_{t\ell} \sim \log \Gam(a_0, b_0)$ is a conditionally conjugate prior;
specifically, we have
\begin{align*}
  \pi_\mu(d\mu) \prod_{X_i \stackrel{\Tree}{\leadsto} \ell} f_\eta(Y_i \mid \lambda_i + \mu) 
  =
  \frac{b_0^{a_0}\exp(\sum_{X_i\leadsto\ell} Y_i \, \lambda_i) \, }{\Gamma(a_0)\prod_{X_i\leadsto\ell} Y_i!}
  \exp\{\mu(a_0 + \sum_{X_i\leadsto\ell} Y_i) - e^\mu (b_0 + \sum_{X_i \leadsto\ell} e^{\lambda_i})\},
\end{align*}
which we recognize as proportional to a $\log \Gam(a_0 + \sum_{i:X_i\leadsto\ell} Y_i, b_0 + \sum_{i:X_i\leadsto\ell} e^{\lambda_i})$ distribution.

\begin{algorithm}[t]
  \caption{One iteration of a generalized Bayesian backfitting algorithm for
    updating $(\Tree_t, \sM_t)$\label{alg:gbb}}
  \textbf{Input:} $\{\Tree_t, \sM_t : t = 1,\ldots,T\}$, $\bY, \bX, \eta, q(\cdot\mid\cdot)$
  \begin{algorithmic}[1]
    \For{$t = 1,\ldots,T$}
    \State Compute $\lambda_i \gets \sum_{k \ne t} g(X_i; \Tree_k, \sM_k)$ for $i = 1,\ldots,N$.
    \State Propose a new tree structure $\Tree' \sim q(\Tree' \mid \Tree_t)$.
    \State Compute the integrated likelihoods $\Lambda(\Tree_t)$ and $\Lambda(\Tree')$ where
    \begin{align*}
      \Lambda(\Tree) = \prod_{\ell \in \Leaves(\Tree)} \int \pi_\mu(\mu) \, \prod_{i : X_i \stackrel{\Tree}{\leadsto} \ell} f_\eta(Y_i \mid \lambda_i + \mu) \ d\mu.
    \end{align*}
    \State Compute the acceptance probability
    \begin{align*}
      A = \min\left\{
      \frac{\Lambda(\Tree') \, \pi_\Tree(\Tree') \, q(\Tree_t \mid \Tree')}
           {\Lambda(\Tree_t) \, \pi_\Tree(\Tree_t) \, q(\Tree' \mid \Tree_t)}
      , 1 \right\}.
    \end{align*}
    \State With probability $A$, set $\Tree_t \gets \Tree'$; otherwise, leave $\Tree_t$ unchanged.
    \State Sample $\sM_t$ from its full conditional distribution.
    \EndFor
  \end{algorithmic}
\end{algorithm}

Unfortunately, for many models of interest it will not be possible to find a $\pi_\mu$ which is conjugate to $\prod_i f_\eta(Y_i \mid \lambda_i + \mu)$. The class of models for which this is feasible is, in fact, surprisingly narrow: for example, one cannot leverage the conjugacy of the beta distribution to the binomial likelihood to construct a generalized Bayesian backfitting algorithm. One possible solution, which was used by \citet{chipman2021mbart} to implement a \emph{monotone} variant of BART, is to compute $\int \pi_\mu(\mu) \prod_{i : X_i \stackrel{\Tree}{\leadsto}\ell} f_\eta(Y_i \mid \lambda_i + \mu) \ d\mu$ numerically and then sample $\sM_t$ using a discrete approximation to the posterior; this introduces new problems, as it requires both approximating the posterior on a grid and evaluating the likelihood at a large number of grid points. In the following section, we show how to bypass the need for conjugacy via RJMCMC.

\section{Implementing Generalized BART with RJMCMC}
\label{sec:rjmcmc}

We now show how to implement the generalized BART model using a generic reversible jump Markov chain Monte Carlo (RJMCMC) algorithm. Because RJMCMC has a reputation for being difficult to implement, and given the breadth of applications we want to consider, it is essential that the algorithms we propose depend on neither tuning parameters nor the details of a given problem. 

We also provide a ``default'' prior for routine use which works well across many problems. This is essential for widespread adoption of our approach, as prior specification is a barrier to the use of Bayesian nonparametric methods by non-experts.

\subsection{Reversible Jump Markov Chain Monte Carlo on Trees}
\label{sec:parameter-expanded}

Throughout this section, we consider updating a regression tree $(\Tree_t, \sM_t)$ with the quantities $\blambda = (\lambda_1,\ldots,\lambda_N)$ and $\eta$ fixed, where $\lambda_i = \sum_{k \ne t} g(X_i; \Tree_k, \sM_k)$. To lighten notation, we will suppress dependence of most quantities in this section on $(\bY, \bX, \eta, \blambda)$, and we will drop the index $t$ from $(\Tree_t, \sM_t)$. Conditional on the $\lambda_i$'s and the nuisance parameter vector $\eta$, the model for the data is $Y_i \sim f_\eta\{y \mid \lambda_i + g(X_i; \Tree, \sM)\}$. The likelihood is then given by
\begin{align}
  \label{eq:Ell}
  \Ell(\Tree, \sM)
  =
  \prod_{\ell \in \Leaves(\Tree)} \prod_{i : X_i \leadsto \ell} f_\eta(Y_i \mid \lambda_i + \mu_{\ell}).
\end{align}
This quantity plays the same role in our RJMCMC scheme as the integrated likelihood $\Lambda(\Tree)$ does in the generalized Bayesian backfitting algorithm of Section~\ref{sec:generalized-bb}.

We consider the following Metropolis-Hastings proposals, which are directly analogous to standard proposals for the Bayesian CART of \citet{chipman1998bayesian}; our proposals operate on $(\Tree, \sM)$ rather than just $\Tree$.

\begin{description}[leftmargin = 0em, style = unboxed]
\item[\Birth]

  Randomly choose a leaf node $\ell \in \Leaves(\Tree)$ and sample a splitting rule $[x_{j_\ell} \le C_{\ell}]$. Convert $\ell$ from a leaf to a branch with two leaf children and
  sample $(\mu'_{\ell L}, \mu'_{\ell R}) \sim G_{\Birth}(\mu'_{\ell L}, \mu'_{\ell R})$ where $G_{\Birth}(\cdot, \cdot)$ is a proposal distribution to be described in Section~\ref{sec:choice}.
  
\item[\Death]

  Randomly choose a branch node $b \in \NOG(\Tree)$ and convert $b$ from a branch to a leaf (deleting its children). Then sample $\mu'_b \sim G_{\Death}(\mu')$ where $G_{\Death}(\cdot)$ is a proposal distribution to be described in Section~\ref{sec:choice}.
  
\item[\Change]

  Randomly choose a branch node $b \in \NOG(\Tree)$ and sample a new splitting rule $[x_{j'_b} \le C'_\ell]$ from the prior. Then sample new leaf node predictions $(\mu'_{bL}, \mu'_{bR}) \sim G_{\Change}(\mu'_{bL}, \mu'_{bR})$ where $G_{\Change}(\cdot,\cdot)$ is a proposal distribution to be described in Section~\ref{sec:choice}.

\end{description}

We now give a valid Metropolis-Hastings acceptance ratio for the \Birth, \Death, and \Change\ moves. It is useful to define, for a given node $n$ (not necessarily a leaf), the quantity
\begin{align}
  \label{eq:scriptf}
  \sF(n \mid \Tree, \mu) = 
  \pi_\mu(\mu) \, 
  \prod_{i: X_i \leadsto n} f_\eta(Y_i \mid \lambda_i + \mu).
\end{align}
Using \eqref{eq:scriptf}, the likelihood \eqref{eq:Ell} is given by $\Ell(\Tree, \sM) = \prod_{\ell \in \Leaves(\Tree)} \sF(\ell \mid \Tree, \mu_\ell) / \pi_\mu(\mu_\ell)$.

\begin{proposition}
  \label{prop:ar}
  Let $p_{\Birth}(\Tree)$ and $p_{\Death}(\Tree)$ denote the probability of proposing \Birth\ and \Death\ moves to modify $\Tree$ respectively and let $|A|$ denote the size of a finite set $A$. For the \Birth, \Death, and \Change\ moves, accepting the proposed change with probability $1 \wedge R$ leaves the posterior invariant, where 
  \begin{align*}
    R_{\Birth}
    &=
    \frac{\rho_d (1 - \rho_{d+1})^2}
    {(1 - \rho_d)}
    \cdot
    \frac{\sF(\ell L \mid \Tree', \mu'_{\ell L}) \, \sF(\ell R \mid \Tree', \mu'_{\ell R})}
    {\sF(\ell \mid \Tree, \mu_{\ell})}
    \cdot
    \frac{p_{\Death}(\Tree') \, |\NOG(\Tree')|^{-1}}
    {p_{\Birth}(\Tree) \, |\Leaves(\Tree)|^{-1}}
    \cdot
    \frac{G_{\Death}(\mu_\ell)}
    {G_{\Birth}(\mu'_{\ell L}, \mu'_{\ell R})}
    \\
    R_{\Death}
    &=
    \frac{(1 - \rho_d)}{\rho_d (1 - \rho_{d+1})^2}
    \cdot
    \frac{\sF(b \mid \Tree', \mu_b')}{\sF(bL \mid \Tree, \mu_{bL}) \, \sF(bR \mid \Tree, \mu_{bR})}
    \cdot
    \frac{p_{\Birth}(\Tree') \, |\Leaves(\Tree)|^{-1}}
    {p_{\Death}(\Tree) \, |\NOG(\Tree')|^{-1}}
    \cdot
    \frac{G_{\Birth}(\mu_{bL}, \mu_{bR})}
    {G_{\Death}(\mu'_b)}
      \quad \text{and} \\
    R_{\Change}
    &= \frac{\sF(bL \mid \Tree', \mu'_{bL}) \, \sF(bR \mid \Tree', \mu'_{bR})}
    {\sF(bL \mid \Tree, \mu_{bL}) \, \sF(bR \mid \Tree, \mu_{bR})}
    \cdot
    \frac{G_{\Change}(\mu_{bL}, \mu_{bR})}{G_{\Change}(\mu'_{bL}, \mu'_{bR})}.
  \end{align*}

\end{proposition}

Proposition~\ref{prop:ar} can be established by applying the results of \citet{green1995reversible} after introducing a suitable dimension-matching transformation. In the Supplementary Material we give a derivation of $R_{\Birth}$
($R_{\Death}$ being the inverse move and $R_{\Change}$ not requiring RJMCMC). Algorithm~\ref{alg:rjmcmc-bart} summarizes the proposed approach.

\begin{algorithm}[t]
  \caption{One iteration of reversible jump Bayesian backfitting\label{alg:rjmcmc-bart}}
  \textbf{Input:} $\bY, \bX, \eta, \{\Tree_t, \sM_t\}_{t=1}^T$
  \begin{algorithmic}[1]
    \State Set $\lambda_i \gets \sum_{t=1}^T g(X_i; \Tree_t, \sM_t)$ for $i = 1,\ldots, N$.
    \For{$i = 1,\ldots,T$}
      \State Set $\lambda_i \gets \lambda_i - g(X_i; \Tree_t, \sM_t)$ for $i = 1,\ldots,N$.
      \State
        Sample $(\Tree', \sM')$ by randomly choosing between the \Birth, \Death, and \Change\ steps.
      \State
        \parbox[t]{.8\textwidth}{Compute the associated acceptance probability from Proposition~\ref{prop:ar} with $(\Tree_t, \sM_t)$ in place of $(\Tree, \sM)$ and accept $(\Tree', \sM')$ with that probability.}
      \State
        Sample $\sM_t$ targeting its full conditional using (say) slice sampling \citep{slicesampling}.
      \State Set $\lambda_i \gets \lambda_i + g(X_i; \Tree_t, \sM_t)$ for $i = 1,\ldots,N$.
    \EndFor
  \end{algorithmic}
\end{algorithm}


\subsection{Choice of the Proposal Distribution}
\label{sec:choice}

The success of Algorithm~\ref{alg:rjmcmc-bart} depends crucially on the quality of the proposal mechanisms $G_{\Death}(\mu_\ell)$, $G_{\Change}(\mu_\ell)$, and $G_{\Birth}(\mu_{\ell L}, \mu_{\ell R})$. As part of the joint proposal for $(\Tree, \sM)$, these proposals are allowed to depend on $\Tree$ as well as $\bY$ and $\blambda$. An effective proposal should be both accurate and applicable to arbitrary models. To meet this need, we choose the proposal to be a $\Normal(m, v^2)$ distribution constructed using the \emph{Laplace approximation} (see, e.g., \citealp{gelman2013bayesian}, Chapter 13), which requires only that we have access to the first and second derivatives of $\log f_\eta(y \mid \lambda)$. Recall that we define $U_\eta(y \mid \lambda) = \frac{\partial}{\partial\lambda} \log f_\eta(y \mid \lambda)$ and $\OFisher_\eta(y \mid \lambda) = -\frac{\partial}{\partial\lambda} U_\eta(y \mid \lambda)$. Then, for example, in the \Birth\ step we propose $\mu'_{\ell L} \sim \Normal(m_{\ell L}, v_{\ell L}^2)$ where
\begin{align}
    \label{eq:laplace}
    \begin{split}
    m_{\ell L} &= \arg \max_{\mu} \sum_{i : X_i \leadsto \ell L} \log f_\eta(Y_i \mid \lambda_i + \mu) + \log \pi_\mu(\mu)
    \qquad\text{and} \\
    v_{\ell L}^{-2} &= \sum_{i: X_i \leadsto {\ell L}} \OFisher_\eta(Y_i \mid \lambda_i + m_{\ell L}) - \frac{d^2}{d\mu^2} \log \pi_\mu(\mu) |_{\mu = m_{\ell L}}.
    \end{split}
\end{align}
The values $m_{\ell L}$ and $v_{\ell L}$ can be computed using, for example, Newton's method: starting from $\mu_{\ell L}$, we perform the update
\begin{align*}
  m_{\ell L} \gets m_{\ell L} + \frac{\sum_i U_\eta(Y_i \mid \lambda_i + m_{\ell L}) + \frac{d}{d\mu} \log \pi_\mu(\mu)|_{\mu=m_{\ell L}}}{\sum_i \OFisher_\eta(Y_i \mid \lambda_i + m_{\ell L}) - \frac{d^2}{d\mu^2} \log \pi_\mu(\mu)|_{\mu = m_{\ell L}}},
\end{align*}
until some stopping criterion is reached. Alternatively, the \emph{Fisher scoring} algorithm replaces $\OFisher_\eta(Y_i \mid \lambda_i + \mu)$ with $\Fisher_\eta(\lambda_i + \mu) = \E\{\OFisher_\eta(Y_i \mid \lambda_i + \mu) \mid \blambda, \eta, \mu\}$; in our experience Fisher scoring tends to be more robust than Newton's method, and we will use Fisher scoring whenever it is feasible. Note also that we do not need to compute \eqref{eq:laplace} exactly, as we just want reasonable Gaussian approximations to the full conditional distributions of the leaf node parameters; any inaccuracies are naturally corrected for by their effect on the Metropolis-Hastings acceptance probability. Algorithm~\ref{alg:newton} gives the Fisher scoring algorithm we used in our illustrations, assuming $\pi_\mu(\mu) = \Normal(\mu \mid 0, \sigma^2_\mu)$; to use Newton's method instead, simply replace $\Fisher_\eta(\lambda)$ with $\OFisher_\eta(Y_i \mid \lambda)$ where appropriate.

\begin{algorithm}[t]
    \caption{Fisher scoring for computing $m_\ell$ and $v_\ell$ for the proposal distribution of $\mu_\ell$ \label{alg:newton}}
    \textbf{Input:} $\ell, \Tree, \bY, \bX, \blambda, \eta, \sigma^2_\mu$

    \textbf{Let:} $U_\eta(\mu, \ell) = \sum_{i: X_i \leadsto \ell} U_\eta(Y_i \mid \lambda_i + \mu) - \mu/\sigma^2_\mu$ and $\Fisher_\eta(\mu,\ell) = \sum_{i: X_i \leadsto \ell} \Fisher(\lambda_i + \mu) + 1/\sigma^2_\mu$.
    \begin{algorithmic}[1]
      \State Initialize $m_\ell$: for \Birth\ moves, use the value of $\mu$ from the parent node; for \Death\ moves use $(\mu_{\ell L} + \mu_{\ell R}) / 2$
      \While{$|U_\eta(m_\ell,\ell)| > \Fisher_\eta(m_\ell,\ell)^{1/2} / 10$}
        \State $m_\ell \gets m_\ell + U_\eta(m_\ell,\ell) / \Fisher_\eta(m_\ell,\ell)$
      \EndWhile
      \State  $v \gets \Fisher_\eta(m_\ell, \ell)^{-1/2}$
      \State \Return $(m_\ell, v_\ell)$
    \end{algorithmic}
\end{algorithm}

Conveniently, the use of a highly informative Gaussian prior for the leaf parameters has benefits for the accuracy of the Laplace approximation. First, the likelihood is encouraged to be nearly Gaussian even if little data is associated to a particular node. Second, because the prior shrinks the $\mu$'s towards zero, Newton's method and Fisher scoring generally converge very quickly even if we initialize the algorithm naively at $\mu = 0$. 

\subsection{Choice of the Prior Distribution}

There are three modeling choices we must make for the prior distribution: the choice of the prior on the tree $\pi_\Tree$, the choice of the prior on the leaf node parameters $\pi_\mu(\mu)$, and the choice of the prior on the nuisance parameter $\eta$. As $\eta$ is obviously problem specific we offer no general guidelines on the selection of its prior. For $\pi_\Tree$ we have found little reason to go beyond the default priors given by \citet{chipman2010bart}, which take $\gamma = 0.95$, $\beta = 2$, and $T \in \{50,200\}$, although there may be some value in choosing $T$ by cross-validation. Additionally, as suggested by \citet{linero2016bayesian}, we recommend replacing the uniform distribution for $j_b$ described in Section~\ref{sec:bart-review} with $j_b \sim \Categorical(s)$ where the hyperparameter $s$ is given a $\Dirichlet(\xi/P,\ldots,\xi/P)$ hyperprior; this allows the model to filter out irrelevant variables much more effectively than the original BART prior.

The choice of $\pi_\mu$ is less straight-forward. We are no longer constrained by the conjugacy requirements of the generalized Bayesian backfitting algorithm, and so for simplicity we take $\pi_\mu(\mu) = \Normal(\mu \mid 0, \sigma^2_\mu)$. We have found, however, that our RJMCMC scheme can be sensitive to the choice of $\sigma^2_\mu$: if a value of $\sigma^2_\mu$ is chosen which is too large, the algorithm gets ``stuck'' early on and does not progress towards the stationary distribution. The appropriate scale for $\sigma^2_\mu$ will typically be problem specific, making it difficult to make a general recommendation. One strategy we have found to work well is to use a half-Cauchy prior $\pi(\sigma_\mu) \propto (1 + \sigma_\mu^2 / c^2)^{-1}$ for some small $c$ (say, $c = k / \sqrt{T}$ where $T$ is the number of trees and $k$ is $1$ or $0.1$). At the beginning of the chain, the small value of $c$ ensures that the chain does not get stuck, while the heavy tails of the Cauchy distribution ensure that --- as the chain approaches the stationary distribution --- the posterior will be able to visit scales of $\sigma_\mu$ larger than $c$. In practice, it may require experimentation to find a value of $c$ for which the chain mixes well.

\section{Illustrations}
\label{sec:applications}

\subsection{Sanity Checks: Semiparametric Regression and Classification with the Logistic Link}

To understand if there are any striking limitations of the RJMCMC approach, we apply it to two problems for which there are existing algorithms: the semiparametric regression problem \eqref{eq:semipar-bart} and nonparametric classification with the logistic link. Going in, we should expect that the RJMCMC algorithm should be inferior in terms of mixing to the algorithm of \citet{chipman2010bart}, as RJMCMC does not use the integrated likelihood (which is available in closed form) to propose changes. For logistic regression it is less clear what to expect, as the algorithm of \citet{sparapani2021nonparametric}, which we compare to, makes use of a data augmentation strategy of \citet{holmes2006bayesian} that itself can substantially slow down mixing. We consider a typical benchmark function for BART methods which takes
\begin{align}
    \label{eq:fried}
    r_F(x) = 10 \sin(\pi  \, x_1  \, x_2) +  20(x_3 - 0.5)^2 + 10 \, x_4 + 5 \, x_5,
\end{align}
with $X_{ij}$ irrelevant for all $j > 5$. 

We compare our RJMCMC algorithm to the methods implemented in the \R\ package \Bart. We choose this package specifically because, to the best of our knowledge, it is the only publicly-available package which implements BART with the logistic link; like our default prior, it also implements the sparsity-inducing Dirichlet hyperprior of \citet{linero2016bayesian}. We remark that \Bart\ differs slightly in how the prior is specified, and for this reason we do not expect that the predictive performance will be precisely the same between the two methods; if RJMCMC performs better, however, this gives us some assurance that the algorithm is correct and mixes well enough to produce reasonable predictions. In all cases we compare RJMCMC and \Bart\ on a single simulated dataset, however the results we present are typical of all replications of the simulations we have performed.

We first consider the semiparametric regression problem \eqref{eq:semipar-bart} with nuisance parameter $\eta = \sigma$ and $r_0(x) = r_F(x)$ where $r_F(x)$ is given by \eqref{eq:fried} with $\sigma^2 = 1$, $N = 500$, and $P = 20$. For both methods we ran the Bayesian backfitting algorithm for $10,000$ iterations, with the first $5,000$ discarded to burn-in. For each iteration, we computed the mean squared error $\MSE = N_{\text{test}}^{-1} \sum_i \{Y^\star_i - r(X_i^\star)\}^2$ where $(Y^\star_i, X^\star_i)$ is a collection of $500$ heldout samples. 

\begin{figure}[t]
    \centering
    \includegraphics[width=.9\textwidth]{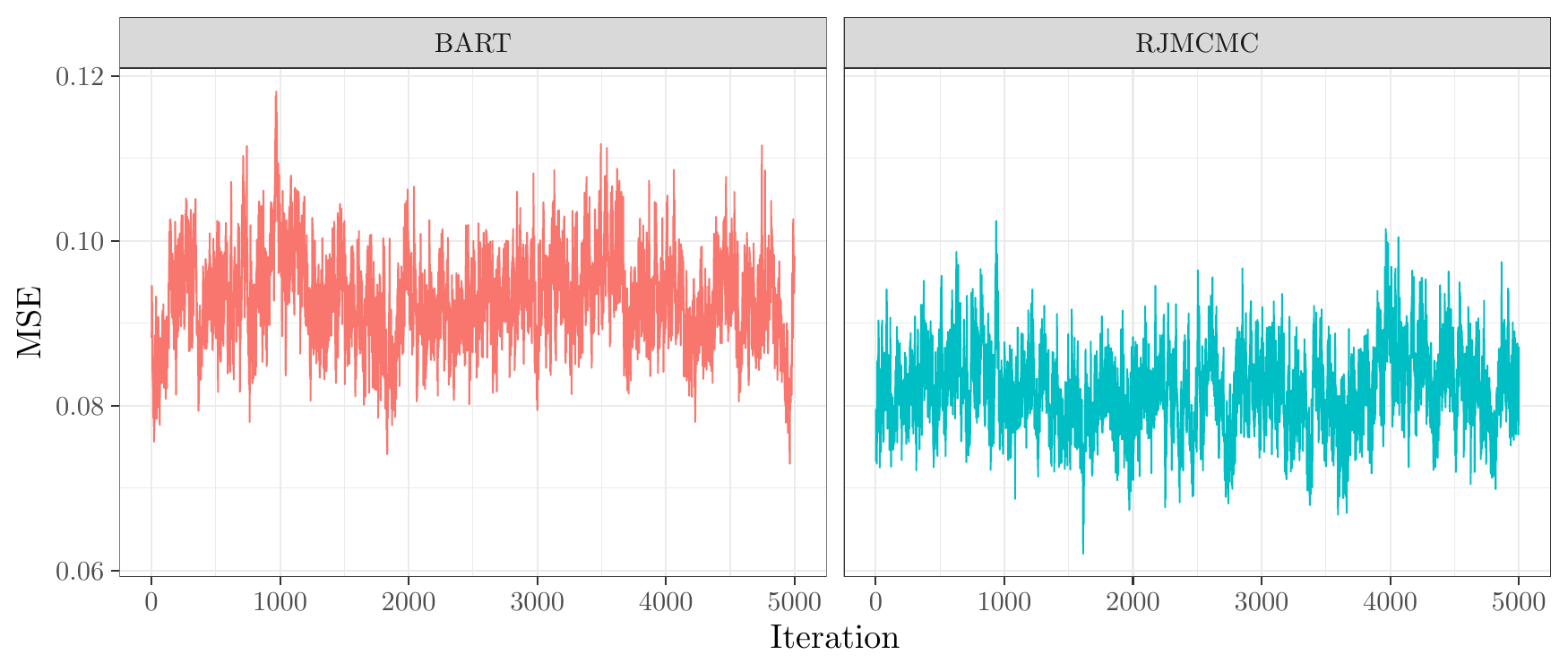}
    \caption{Traceplot of the heldout mean squared error for the semiparametric regression model. Left: results as implemented in the \texttt{BART} package. Right: results using our RJMCMC algorithm.}
    \label{fig:batman_compare}
\end{figure}

Figure~\ref{fig:batman_compare} displays the samples of $\MSE$ for both approaches. Both methods are similar in terms of mixing; in particular, the mixing of RJMCMC does not appear to be appreciably worse. We also see that RJMCMC results in a lower $\MSE$ on average.

For the classification problem we take
$Y_i \sim \Bernoulli[\sigmoid\{r_0(X_i)\}]$ 
where $\eta = \emptyset$, $r_0(x) = \frac{r_F(x) - 14}{5}$, and $\sigmoid(x) = (1 + e^{-x})^{-1}$ is the logistic function; this normalization of $r_F(x)$ was chosen so that $r_0(X_i)$ has approximately mean $0$ and variance $1$. 

Data augmentation can be applied to fit BART classification models using the logistic link. In fact, there are at least three approaches to this: the scale-mixtures-of-normals approach of \citet{holmes2006bayesian}, the \Polya-gamma approach of \citet{polson2013bayesian}, and the gamma augmentation approach of \citet{murray2020log}. The downside of these approaches is that data augmentation can slow down mixing substantially, especially in cases where the outcome distribution is highly imbalanced \citep{johndrow2018mcmc}.

\begin{figure}[t]
    \centering
    \includegraphics[width=.9\textwidth]{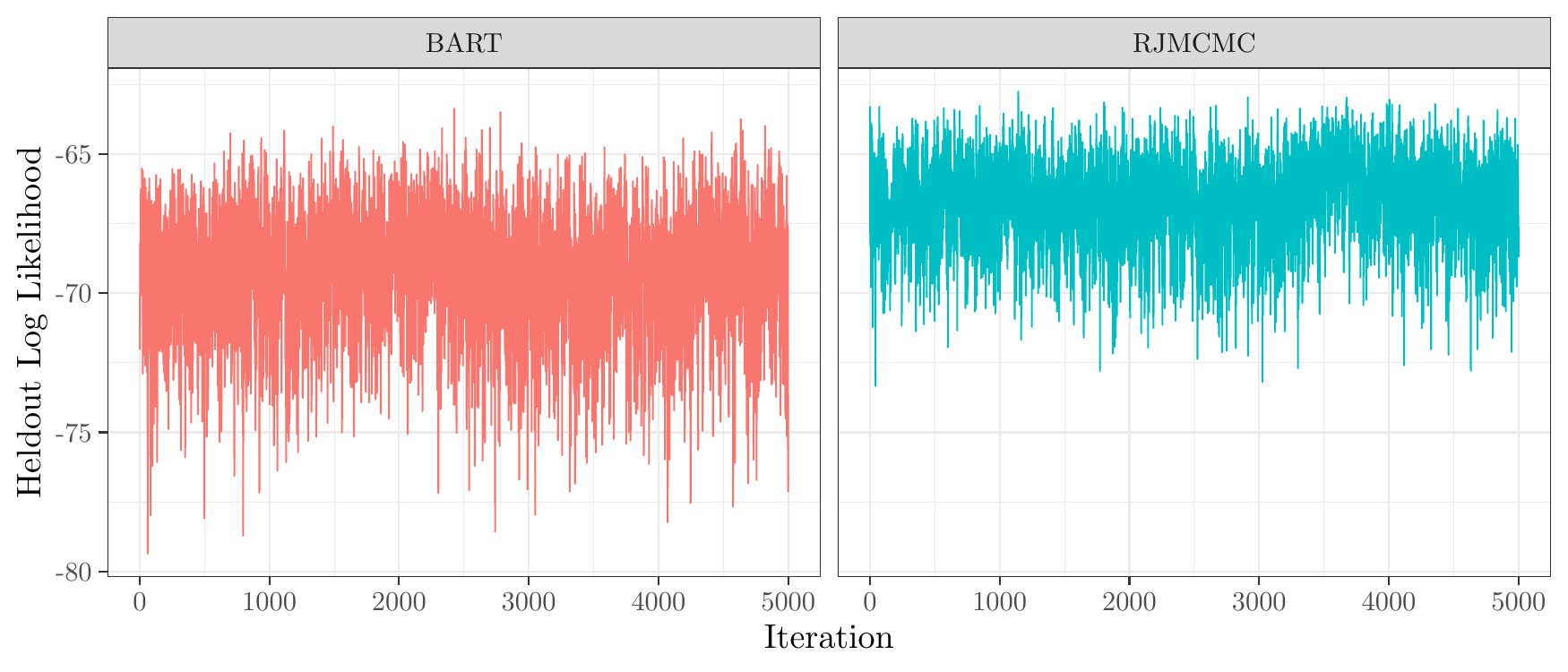}
    \caption{Traceplot of the heldout log-likelihood for the logistic link BART model. Left: results using the \texttt{BART} package. Right: results using our RJMCMC algorithm.}
    \label{fig:testing}
\end{figure}

Our RJMCMC algorithm removes the need for data augmentation entirely, and requires only that we plug in the likelihood, score, and Fisher information given by
\begin{align*}
    \log f(y \mid \lambda) &= y \log \sigmoid(\lambda) + (1 - y) \log\{1 - \sigmoid(\lambda)\}, \\
    U(y \mid \lambda) &= y - \sigmoid(\lambda), \qquad \text{and}\qquad \text{and}  \\
    \Fisher(\lambda) &= \sigmoid(\lambda) \, \{1 - \sigmoid(\lambda)\}. 
\end{align*}
We fit the classification model using both the \Bart\ package (which uses the data augmentation scheme of \citealp{holmes2006bayesian}) and our RJMCMC algorithm. For each iteration we record the heldout log-likelihood
\begin{math}
   \sum_i Y_i^\star \log \sigmoid\{r(X_i^\star)\} + (1 - Y_i^\star) \log[1 - \sigmoid\{r(X_i^\star)\}],
\end{math}
where $(Y_i^\star, X_i^\star)$ are $500$ heldout observations. In Figure~\ref{fig:testing} we give traceplots of the heldout log-likelihood for both methods, and we again observe that RJMCMC does not mix appreciably worse than \Bart\ while producing better predictions on the heldout data.

\subsection{Variance Modeling}

We now turn our attention to generalized BART models that cannot be fit with existing Bayesian backfitting algorithms. A common concern when constructing a regression model is \emph{heteroskedasticity} of the error distribution. A selling point of generalized linear models, for example, is that they handle the mean-variance relationships inherent to proportion or count data.

In this section we consider BART models which allow for a specified (but essentially arbitrary) mean-variance relationship using a Gaussian working model. Specifically, we set
\begin{align}
  \label{eq:varmod}
  [Y_i \mid X_i] \sim \Normal\{m_i, \phi \, V(m_i)\}
\end{align}
where $m_i = g\{r(X_i)\}$. Here, $g(\mu)$ and $V(m)$ are user-specified functions which relate $r(x)$, the mean, and the variance. In this case, $\eta = (\phi, V(\cdot))$. In the Supplementary Material we show that $U_\eta(y \mid \lambda)$  and $\Fisher_\eta(\lambda)$ are given by
\begin{align*}
  U_\eta(y \mid \lambda)
  &= \left(
    -\frac{V'(m)}{2V(m)} + \frac{V'(m) (y - m)^2}{2\phi V(m)^2}
    + \frac{y - m}{\phi V(m)}
    \right) g'(\lambda)
  \\
  \Fisher_\eta(\lambda)
  &= 
    \left( \frac{V'(m)^2}{2 V(m)^2} + \frac{1}{\phi V(m)} \right) g'(\lambda)^2
\end{align*}
where $m = g(\lambda)$, $V'(m) = \frac{d}{dm} V(m)$, and $g'(\lambda) = \frac{d}{d\lambda} g(\lambda)$. Additionally, the full conditional of $\tau = \phi^{-1}$ is $\Gam\{\tau \mid N/2, \nicefrac{1}{2}\sum_i (Y_i - m_i)^2 / V(m_i)\} \times \pi(\tau)$.
Plugging these expressions into our generic RJMCMC scheme, we can fit a BART model to any mean-variance relationship.

\begin{figure}[t]
  \centering
  \includegraphics[width=.9\textwidth]{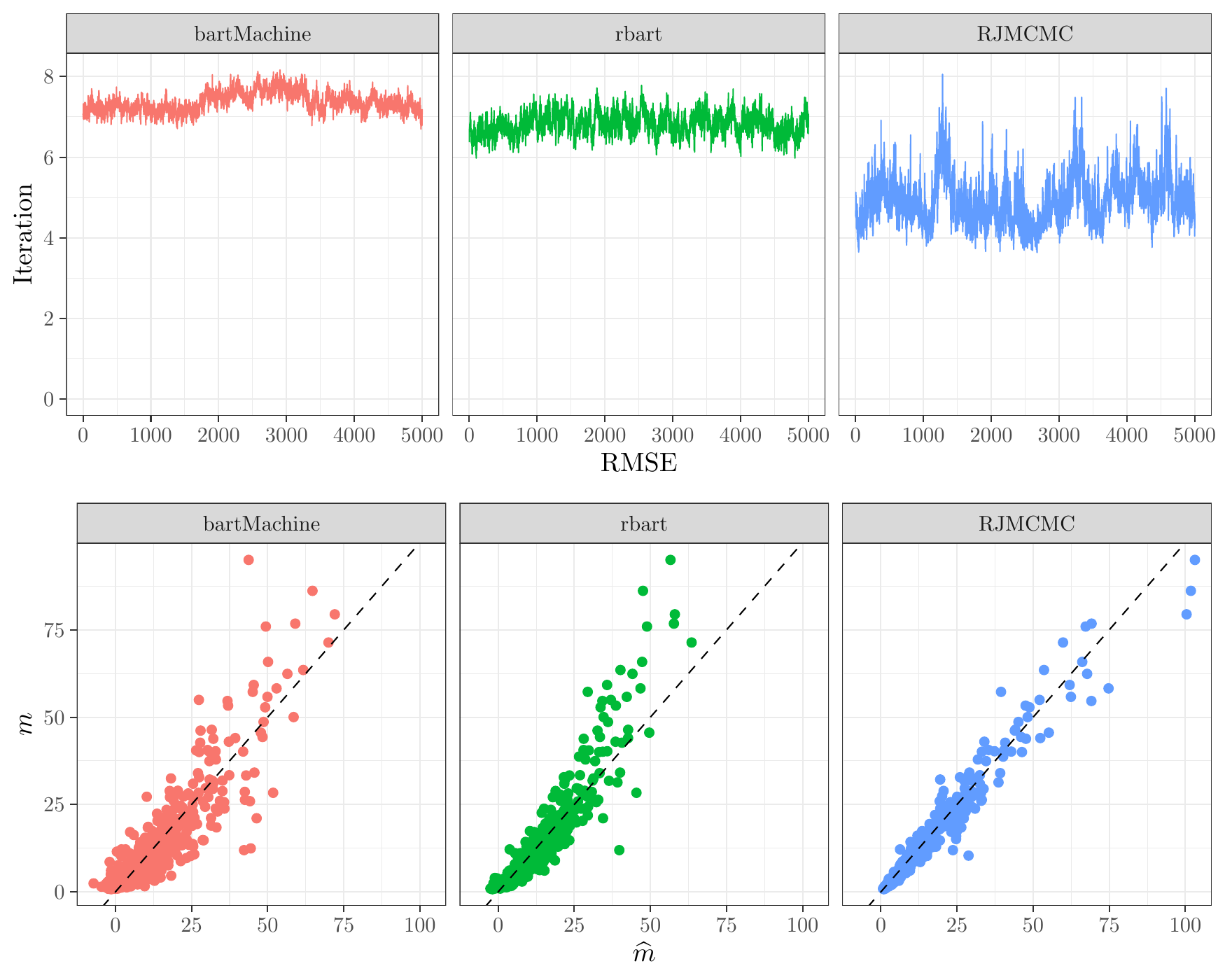}
  \caption{Top: traceplot of RMSE on heldout samples for each method. Bottom: Plot of the Bayes point estimate of $\E(Y_i \mid X_i)$ $(\widehat m_i)$ against its true value ($m$); the line $m = \widehat m$ is given by the dashed line.}
  \label{fig:testing_var}
\end{figure}

To illustrate, we generated $Y_i \sim \Poisson(m_i)$ with $m_i = \exp\{r(X_i)\}$, which implies the mean-variance relation $V(m) = m$. We took
\begin{math}
  m_i = \exp\left\{2 + \left( \frac{r_F(X_i) - 14}{5} \right)\right\}
\end{math}
with $r_F(x)$ given by \eqref{eq:fried}. We compare the following three BART implementations.
\begin{itemize}
\item \textbf{bartMachine}:
  A standard BART model which takes $Y_i \sim \Normal\{r(X_i), \sigma^2\}$, fit using the \bartMachine\ package.
\item \textbf{rbart}:
  A heteroskedastic BART model of \citet{pratola2017heteroscedastic}, which takes $Y_i \sim \Normal\{r(X_i), \sigma^2(X_i)\}$. This model was fit using the \rbart\ package. 
\item \textbf{RJMCMC}:
  The BART model \eqref{eq:varmod} which takes $Y_i \sim \Normal\{e^{r(X_i)}, \phi e^{r(X_i)}\}$.
\end{itemize}
The goal of this comparison is to determine (i) if our RJMCMC algorithm is capable of fitting \eqref{eq:varmod}, (ii) if there is a substantial gain in performance from modeling the variance, and (iii) if there is additional gain from correctly specifying the mean-variance relationship. We compare methods based on the root mean-squared error $\RMSE = \sqrt{N_{\text{test}}^{-1} \sum_i (m^\star_i - \widehat m^\star_i)^2}$ on a collection of 500 heldout samples $(X_i^\star, Y_i^\star)$. Results are given in Figure~\ref{fig:testing_var}. We see from the traceplot of RMSE that the RJMCMC model results in a substantially lower RMSE on heldout data, and that the RMSE mixes well for all three methods; while \rbart\ is able to account for heteroskedasticity, it gives only a modest improvement over \bartMachine. For both \bartMachine\ and \rbart\ we see that the models tend to underestimate $m_i$ when $m_i$ is large. The overall RMSEs using the Bayes estimate for each method are 6.67 (\bartMachine), 5.71 (\rbart), and 3.25 (RJMCMC).

\subsection{Accelerated Failure Time Models}
\label{sec:aft}

We now illustrate our approach on several \emph{accelerated failure time} (AFT) models for survival analysis \citep{wei1992accelerated}. Let $T_i$ denote a survival time
and let $C_i$ denote the censoring time such that we observe $Y_i = \min\{T_i, C_i\}$ and $\delta_i = I(Y_i = T_i)$. The accelerated failure time model takes
\begin{align}
  \label{eq:aft}
  \log T_i = r(X_i) + \sigma \, \epsilon_i
\end{align}
where $\epsilon_i$ belongs to some parametric family of distributions; common choices include the normal, logistic, and log-gamma distributions. The log-likelihood of the AFT model is given by
\begin{align*}
  \Ell(r, \eta) = \prod_i S_\epsilon\left( \frac{\log Y_i - r(X_i)}{\sigma} \right)
  \left\{ \frac{h_\epsilon\left( \frac{\log Y_i - r(X_i)}{\sigma} \right)}{\sigma} \right\}^{\delta_i},
\end{align*}
where $S_\epsilon(t)$ is the survival function of $\epsilon_i$, $f_\epsilon(t)$ is the density of $\epsilon_i$, and $h_\epsilon(t) = f_\epsilon(t) / S_\epsilon(t)$ is the hazard function of $\epsilon_i$. We consider $\epsilon_i \sim \Logistic(0,1)$ and $\epsilon_i \sim \log\Gam(\alpha,\alpha)$. These models correspond to \emph{log-logistic} $(\eta = \sigma)$ and \emph{generalized gamma} $(\eta = (\sigma, \alpha))$ AFT models for $T_i$ respectively. For both models, we consider a ground truth of $r_0(x) = r_F(x)$ and $\sigma = 1$.

The log-logistic model, for which $f_\epsilon(t) = \sigmoid(t) \, \{1 - \sigmoid(t)\}$, is particularly convenient in that both the survival function $S_\epsilon(t) = 1 - \sigmoid(t)$ and hazard function $h_\epsilon(t) = \sigmoid(t)$ can be written in closed form.
To this point, the generalized gamma model is the first model for which we cannot compute
$\Fisher_\eta(\lambda)$
in closed form. We therefore use this as an opportunity to show that our methodology works well even when we approximate the required derivatives numerically. Given a function $w(\mu)$, we use finite differences to approximate the first and second derivatives as $w'(\mu) \approx \frac{w(\mu + \Delta) - w(\mu - \Delta)}{2\Delta}$ and $w''(\mu) \approx \frac{w(\mu + \Delta) - 2 w(\mu) + w(\mu - \Delta)}{\Delta^2}$ with $\Delta = 10^{-6}$.

\begin{figure}
  \centering
  \includegraphics[width=1\textwidth]{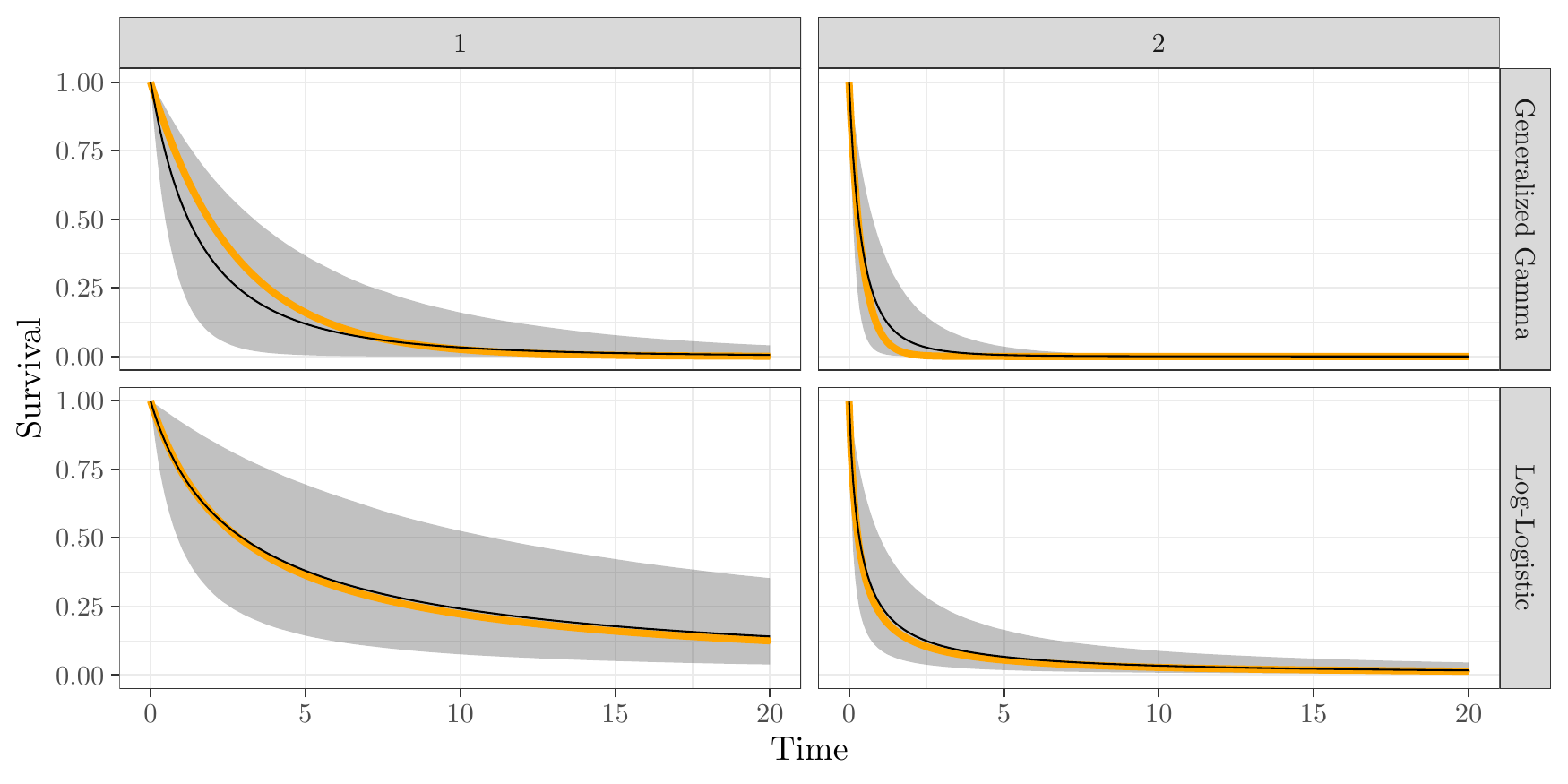}
  \caption{Estimated (black) and true (orange) survival curves, with 95\% posterior credible bands, for two randomly selected observations for the generalized gamma and log-logistic AFT models.}
  \label{fig:bands}
\end{figure}

We simulate data from both models with $r_0(x) = \frac{r_F(x) - 14}{5}$ and $(N,P) = (500,10)$. We censored the data at randomly by sampling $C_i$ and $T_i$ from the same distribution; by design, this results in roughly 50\% of the samples being censored regardless of the value of $X_i$. For the generalized gamma model, we used the ground truth $\sigma_0 = \alpha_0 = 1$. 

Overall, we found that both chains mixed well, with the exception that the mixing of $\sigma$ and $\alpha$ was poor for the generalized gamma AFT model; this poor mixing occurs because $\sigma$ and $\alpha$ are highly correlated in the posterior, and should be updated jointly rather than with the slice sampler we used. We also found that the log-logistic model took less time per iteration because the survival function of the log-logistic model is available in closed form. In general, $\sigma$ and $\alpha$ are poorly identified due to the fact that both parameters are largely variance parameters for $\log T_i$, with 95\% credible intervals being $\sigma \in (0.67, 3.72)$ and $\alpha \in (0.60, 7.45)$. Despite this, the chain mixes very well on the variance parameter $V = \Var(\log T_i \mid r, \sigma, \alpha) = \sigma^2 \, \psi'(\alpha)$, with the Bayes estimate $\widehat V = 1.72$ being very close to the true value $V_0 = 1.64$.

Plots like those in Figure~\ref{fig:testing_var_gamma} (right) and Figure~\ref{fig:testing_var} (bottom) are given in the Supplementary material; they show that both the log-logistic and generalized gamma models recover $r(x)$ effectively. Estimates of the survival curve, along with 95\% credible bands, for some randomly-sampled observations in a heldout test set are given in Figure~\ref{fig:bands}. We see that the point estimates and credible bands provide accurate inference for the true survival curves

\subsubsection*{Application to Liver Disease Data}

We apply the AFT log-logistic (AFTLL) and generalized gamma (AFTGG) models to a dataset from a randomized clinical trial on time to death for individuals suffering from primary biliary cirrhosis; this data is publicly available as the \texttt{pbc} dataset in the package \texttt{randomForestSRC}. Our goal is to determine which of the parametric families provides the best description of this data. In addition to these models, we consider a semiparametric Weibull model with hazard function of the form
\begin{align*}
  h(t \mid \lambda, k) = \frac{k}{e^\lambda} \left( \frac{t}{e^{\lambda}} \right)^{k-1},
\end{align*}
with the survival time modeled as $T_i \sim h\{t \mid r(X_i), k\}$ and $\eta = k$; a similar model is proposed by \citet{linero2021bayesian}. 

This Weibull model, which sits at the intersection of AFT and proportional hazards models, is a special case of the generalized gamma model with $\alpha = 1$. Additionally, the generalized gamma model includes the log-normal AFT model as a limiting case as $\alpha \to \infty$; this makes the generalized gamma model a potentially useful tool for deciding between different parametric families.

To gain insight into whether different models lead to different qualitative prognoses for patients, we compare the estimates of $r(X_i)$ for the different models in the Supplementary Material. We found that the models agreed remarkably well in their estimates of $r(X_i)$. 

Conversely, we also found that the data did not distinguish well between the different models, particularly for large survival times. In the Supplementary Material, we plot the posterior distribution of the shape parameter $\alpha$ in the generalized gamma model under a $\Uniform(0,40)$ prior, and find that the data is consistent with both the Weibull model $(\alpha = 1)$ and log-normal model $(\alpha \to \infty)$. These models make quite different predictions for the hazard at later timepoints, with the Weibull model having a monotonically-increasing hazard $(k \approx 1.3)$ and the log-normal and log-logistic models both having non-monotone hazards.


Finally, we evaluate the goodness of fit of the AFTLL, AFTGG, and Weibull models using the log-pseudo marginal likelihood (LPML) given by 
\begin{math}
  \sum_i \log f(Y_i, \delta_i \mid \bY_{-i}, \bdelta_{-i}, \bX)
\end{math}
where $\bY_{-i}$ and $\bdelta_{-i}$ denote the vector of event times and censoring indicators with observation $i$ removed, while $\bX = (X_1,\ldots,X_N)$ and $f(Y_i, \delta_i \mid \bY_{-i}, \bdelta_{-i}, \bX)$ is the \emph{predictive density} given by
$\int f_\eta(Y_i, \delta_i \mid r(X_i)) \, \pi(r, \eta \mid \bY_{-i}, \bdelta_{-i}, \bX) \ dr \ d\eta$. The LPML can be conveniently computed from the output of the MCMC sampler using the \texttt{loo} package in \texttt{R}. The fits of all three models are quite similar, with the estimated LPMLs being $(-350.6, -353.4, -350.1)$ for the Weibull, AFTGG, and AFTLL models, respectively. According to LPML, there is a slight preference for the log-normal model, which has a non-monotone hazard, although the Weibull model performs very similarly. This observation is consistent with our findings in the Supplementary Material, where we find that the posterior distribution of the AFTGG model is consistent with both the Weibull ($\alpha = 1$) and log-normal $(\alpha \to \infty)$ models.

\subsection{Gamma Shape Regression}
\label{sec:gamma-shape-regression}

An interesting extension of the accelerated failure time models discussed in Section~\ref{sec:aft} is to allow for the shape of the hazard function itself to depend on the covariates; this would allow some individuals to have monotonically increasing, decreasing, or non-monotone hazards depending on their covariates. One approach to doing this is to model the shape parameter $\alpha$ in the generalized gamma model in a covariate-dependent fashion as well. Towards this end, we consider a gamma regression model which takes $Y_i \sim \Gam\{\alpha(X_i), \beta\}$ (where $\eta = \beta$). A BART model for the related gamma regression model $Y_i \sim \Gam\{\alpha, \beta(X_i)\}$ was considered by \citet{linero2018shared}, who showed that this model can be made conditionally conjugate; due to the fact that $\beta(X_i)$ is not a shape parameter for the gamma distribution, however, this model is not appropriate for modeling changes in the shape of the hazard.


We model the shape parameter on the log scale, taking $\alpha(X_i) = \exp\{r(X_i)\}$. It is then straight-forward to show that 
\begin{align*}
  U_\eta(y \mid \lambda) = e^\lambda \{\log \beta - \psi(e^\lambda) + \log y\}
  \quad \text{and} \quad
  \Fisher_\eta(\lambda) = e^{2\lambda} \, \psi'(e^\lambda)
\end{align*}
where $\psi(\alpha) = \frac{d}{d\alpha} \log \Gamma(\alpha)$ and $\psi'(\alpha) = \frac{d}{d\alpha} \psi(\alpha)$ are the digamma and trigamma functions, respectively.

We simulate data from the model with $(N, \beta) = (100, 1)$ and $r(x) = 2 + \frac{r_F(x) - 14}{5}$ so that $\log \alpha(X_i)$ has roughly mean $2$ and variance $1$, and fit the model with the default prior. Mixing of the RJMCMC scheme is given in the Supplementary Material; summarizing, we found that the chain mixed well. As shown in the right panel of Figure~\ref{fig:testing_var_gamma}, generalized BART is able to accurately recover $r_0(x)$ on a set of heldout covariates $(X_1^\star, \ldots, X_{N_{\text{test}}}^\star)$.


\subsection{Comparison with Tree Boosting}

We now give a brief comparison of the generalized BART model with tree boosting as implemented in the \blackboost\ function in the \mboost\ package in \R. Our comparisons are biased \emph{in favor} of the \mboost\ package: for each comparison, we chose the \mboost\ hyperparameters (the shrinkage parameter \texttt{nu} and the number of boosting iterations \texttt{mstop}) to minimize the error on the test set, and chose the maximal depth of the tree (\texttt{maxdepth}) equal to $2$ to ensure that \mboost\ does not include any spurious higher-order interactions. By contrast, the hyperparameters for generalized BART are either fixed a-priori or learned from the training data.

We compare \blackboost\ to generalized BART on the logistic regression problem, the log-logistic accelerated failure time (AFT) problem, the gamma shape regression problem, and the structure heteroskedastic regression problem. The \mboost\ package implements logistic regression and log-logistic AFT models, and we used the functionality within \mboost\ to build custom procedures for the gamma and heteroskedastic regression models. In each case, accuracy is measured through the mean squared error $\sqrt{N_{\text{test}}^{-1} \sum_i \{r_0(X_i^\star) - \widehat{r}(X_i^\star)\}^2}$ where $(X^\star_1, \ldots, X^\star_{N_{\text{test}}})$ denotes a heldout test set of 500 points and $\widehat r(x)$ denotes the point estimate of $r_0(x)$ (for boosting) or the posterior mean of $r(x)$ (for BART).

\begin{figure}[t]
  \centering
  \includegraphics[width=1\textwidth]{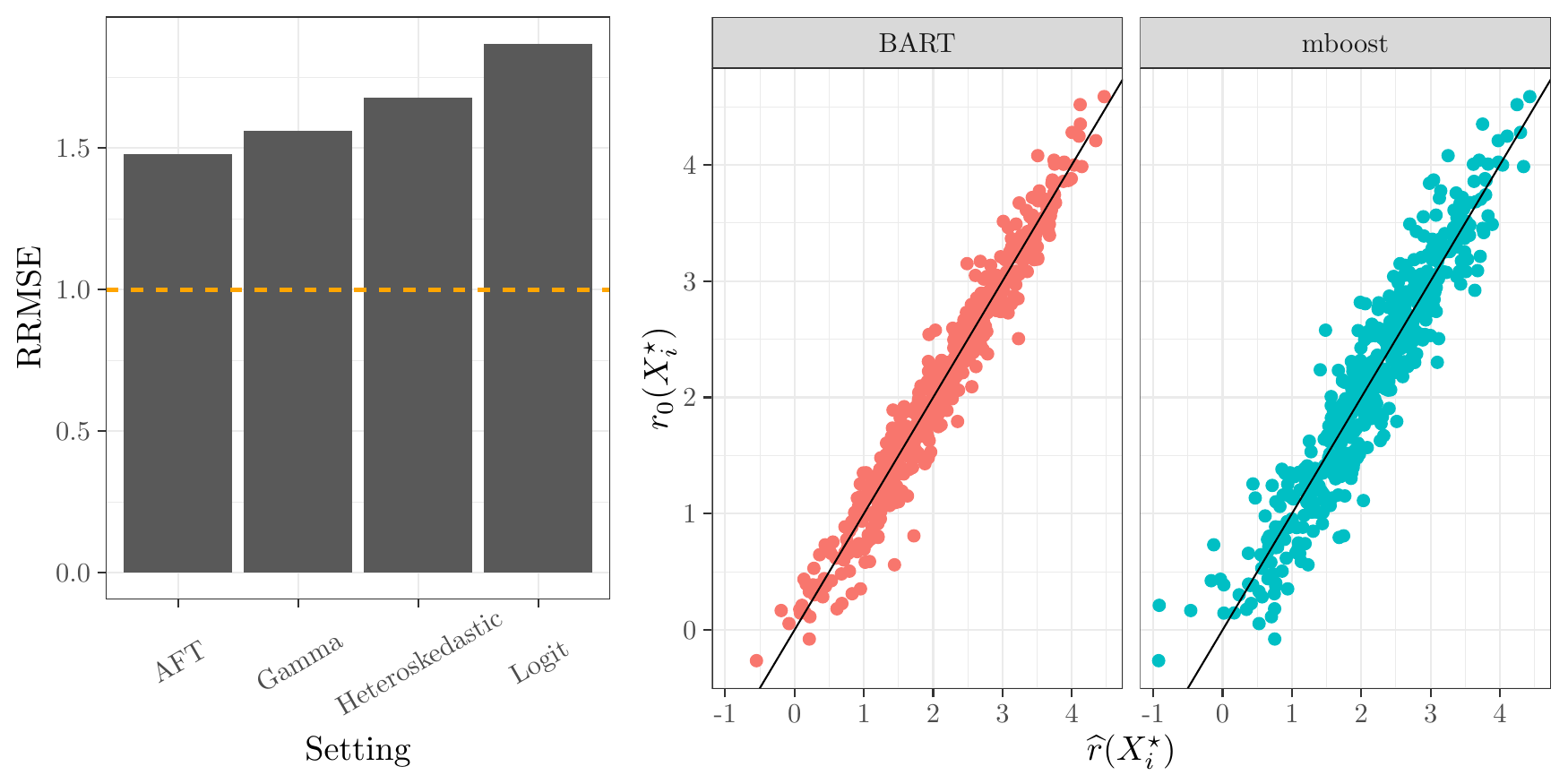}
  \caption{Left: root mean-squared error of optimally-tuned \blackboost\ relative to the root mean-squared error (RRMSE) of the default generalized BART prior for the logistic AFT (AFT), gamma shape (Gamma), structured variance (Heteroskedastic), and logistic regression (Logit) models. To aide visualization, the orange dashed line at $1$ would denote a tie in performance with generalized BART. Right: plots of $\widehat r(X_i^\star)$ against $r_0(X_i^\star)$ for generalized BART and \blackboost\ for the gamma shape regression model.}
  \label{fig:testing_var_gamma}
\end{figure}

Results are given in Figure~\ref{fig:testing_var_gamma}, with $N = 500$ and $P = 10$; the results presented here are representative of what occurs in repeated simulations and are consistent with what occurs for similar simulation experiments \citep{linero2016bayesian}. Despite the simulation settings here being generally favorable to boosting (the hyperparameters were optimally tuned to the test set, there is relatively little noise, and the number of nuisance predictors is small) the results are strongly in favor of generalized BART. Specifically, the RMSE of \blackboost\ ranges from 50\% larger to 80\% larger than the RMSE of generalized BART.

The right panel of Figure~\ref{fig:testing_var_gamma}, which focuses on the gamma shape regression problem, displays $r_0(X_i^\star)$ against $\widehat r(X_i^\star)$ for \blackboost\ and generalized BART, and provides a sanity check that both methods are working as intended. Both sets of predictions cluster around the 45 degree line, with \blackboost\ being less precise.

\section{Discussion}
\label{sec:discussion}

The approach outlined in this article greatly expands the problems to which BART can be applied, and we emphasize that none of the models we applied BART to required any modifications to our algorithm. There are many directions for extending this framework in future work. For example, by modifying the approach to allow for more than one forest \citep{pratola2017heteroscedastic}, we could develop flexible gamma regression models with $Y_i \sim  \Gam\{\alpha(X_i), \alpha(X_i) / \mu(X_i)\}$ or beta  regression models with $Y_i \sim \Beta\{\mu(X_i) \, \phi(X_i), \phi(X_i) - \phi(X_i) \, \mu(X_i)\}$. This could be done using either separate forests, in which case  our methodology extends directly, or using the shared forests approach of \citet{linero2018shared}. The shared forests approach is likely more difficult to implement due to the need for a multivariate Metropolis-Hastings proposal for the reversible jump move.

An additional application of our RJMCMC algorithm is that it can be extended to the \emph{soft} BART models of \citet{linero2017abayesian}. These models --- which have better theoretical and empirical properties than standard BART models when the underlying function $r_0(x)$ is smooth --- can only use conjugate updates for the model \eqref{eq:semipar-bart} to the best of our knowledge.

For the generalized gamma model, we crudely avoided computing the score and Fisher information by using numerical differentiation; this approximates the likelihood, score, and Fisher information using a total of three likelihood evaluations, and so is relatively efficient. We note that it is, in principle, possible to eliminate the need for the user to explicitly compute the derivatives of the likelihood by using software that performs automatic differentiation such as \Tensorflow.

A lingering advantage of gradient boosting over BART is that gradient boosting is much faster and scales better to large datasets. Recently, \citet{he2019accelerated} and \citet{he2021stochastic} substantially closed this gap with their XBART algorithm; however, this approach also requires the same sort of conditional conjugacy as the generalized BART model. It is worth exploring whether our RJMCMC algorithm might be combined with XBART, either to be used after a ``warm-start'' with XBART or to be used to construct a replacement for the XBART splitting criterion.

\bibliographystyle{apalike}
\bibliography{mybib}



\section*{Supplementary material}

# Supplementary Material to Generalized Bayesian Additive Regression Trees: Beyond Conditional Conjugacy

Antonio R. Linero*

## S.1 Additional Illustration Figures

**Mixing for Gamma Shape Regression**

In Figure S.1 we provide mixing of the quantity $\text{RMSE} = \sqrt{N_{\text{test}}^{-1} \sum_i \{r(X_i^\star) - r_0(X_i^\star)\}^2}$ for four independent chains fit to data simulated under the gamma shape regression model described in Section 4.4. Mixing on this quantity, as well as other identified quantities (not provided) are quite good.

**Accelerated Failure Time Fit**

In Figure S.2 we display the predicted values from the generalized BART model against the true values on the heldout test set. We see that both methods recover $r_0(x)$ reasonably well, although the generalized gamma model appears to perform less well for larger predicted values.

**PBC Data**

Figure S.3 displays the estimated baseline hazard function for the AFTLL, AFTGG, and Weibull regression model fits to the `pbc` dataset. Because the errors in the AFT regression

---

*Department of Statistics and Data Sciences, University of Texas at Austin, email: antonio.linero@austin.utexas.edu

1

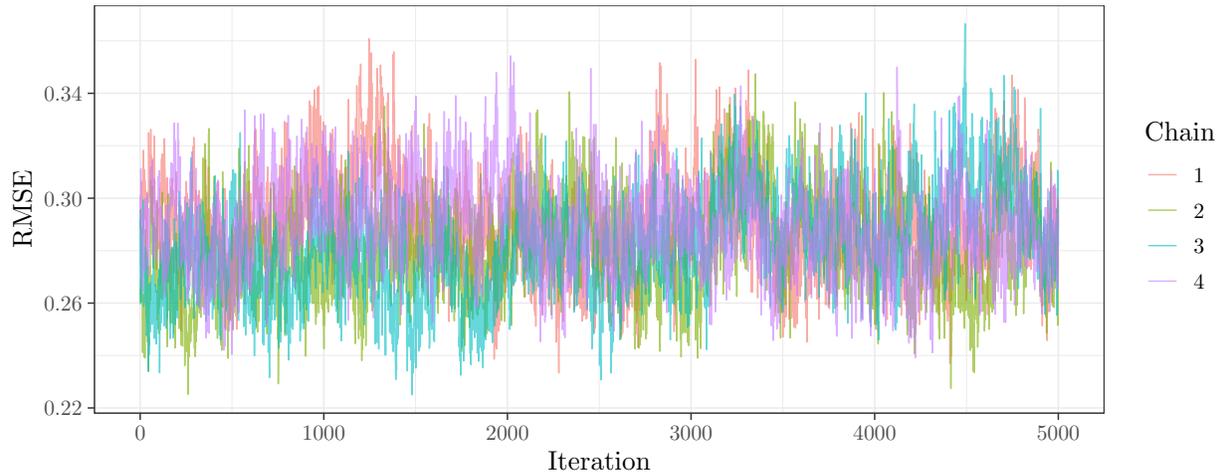

Figure S.1: Traceplots of RMSE for four independent chains fit to data simulated under the gamma shape regression model.

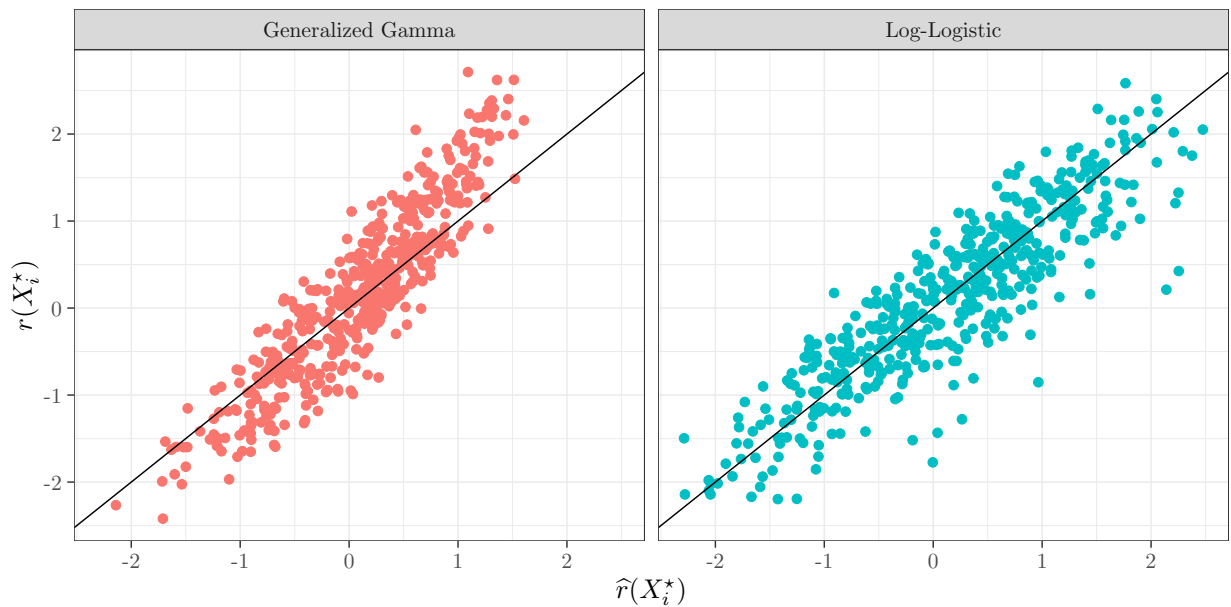

Figure S.2: Plots of the Bayes estimate $\widehat{r}(X_i^\star)$ against $r_0(X_i^\star)$; the line $r_0 = \widehat{r}$ is given by the solid line.

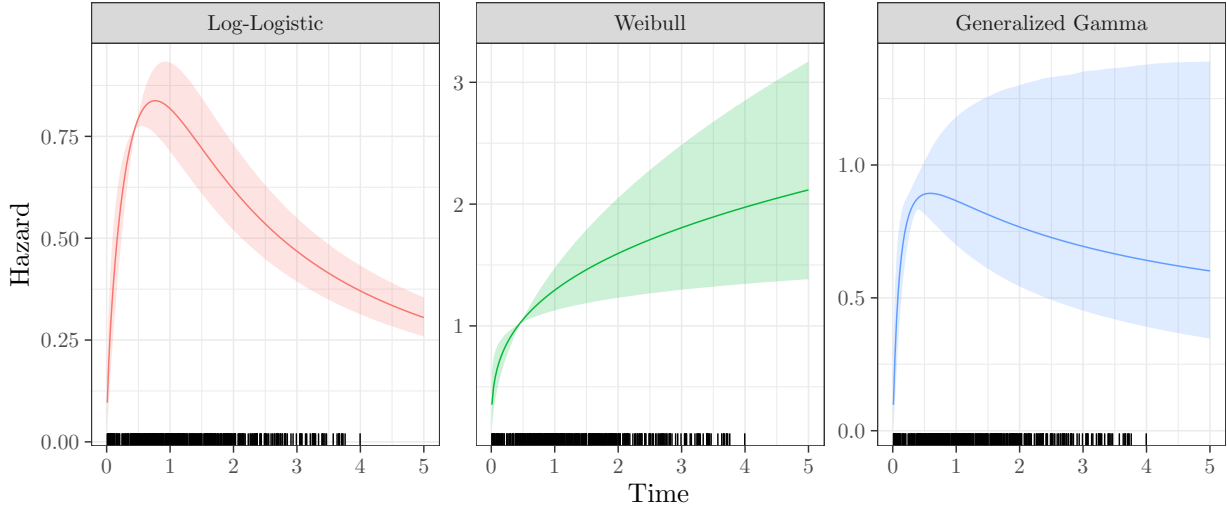

Figure S.3: Bayes estimates of the baseline hazard (i.e., the hazard associated to setting $r(X_i) = 0$) along with 95% credible bands. Event times are given as a rug at the bottom of each panel.

model $\log T_i = r(X_i) + \epsilon_i$ are not necessarily mean 0, we focus on qualitative comparisons of the shapes of the hazards. The best-fitting model (the log-logistic) estimates the hazard to be non-monotone, with a mode relatively early on in the study; conversely, the Weibull model has a monotonically-increasing hazard function. The generalized gamma model, which interpolates between a Weibull ($\alpha = 1$) and log-normal ($\alpha = \infty$) model, adequately reflects the uncertainty of the shape of the hazard for this data — the posterior distribution of $\alpha$ is quite diffuse, and the data is consistent with both the Weibull and log-normal models.

Figure S.4 displays the posterior distribution of $\alpha$ for the generalized gamma model, corroborating the observations above about the relative plausibility of the Weibull and log-normal models. We note that a Uniform$(0, 40)$ prior was used for $\alpha$, explaining the sharp cutoff of the posterior at 40.

Figure S.5 displays the Bayes estimates of $r(X_i)$ for the generalized gamma model against the same for the log-logistic and Weibull models. The agreement in estimates demonstrates both (i) that the models largely agree about the prognostic implications of the covariates and (ii) that the RJMCMC scheme is reliable for all three models.

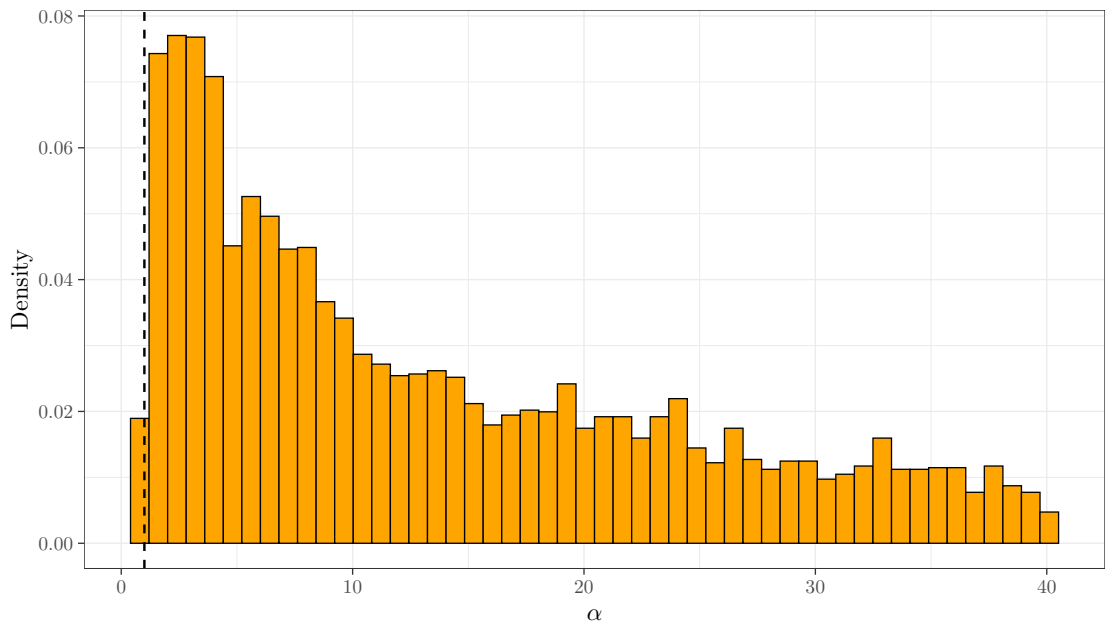

Figure S.4: Posterior of the shape parameter $\alpha$ for the `pbc` data under a generalized gamma model with prior $\alpha \sim \text{Uniform}(0, 40)$.

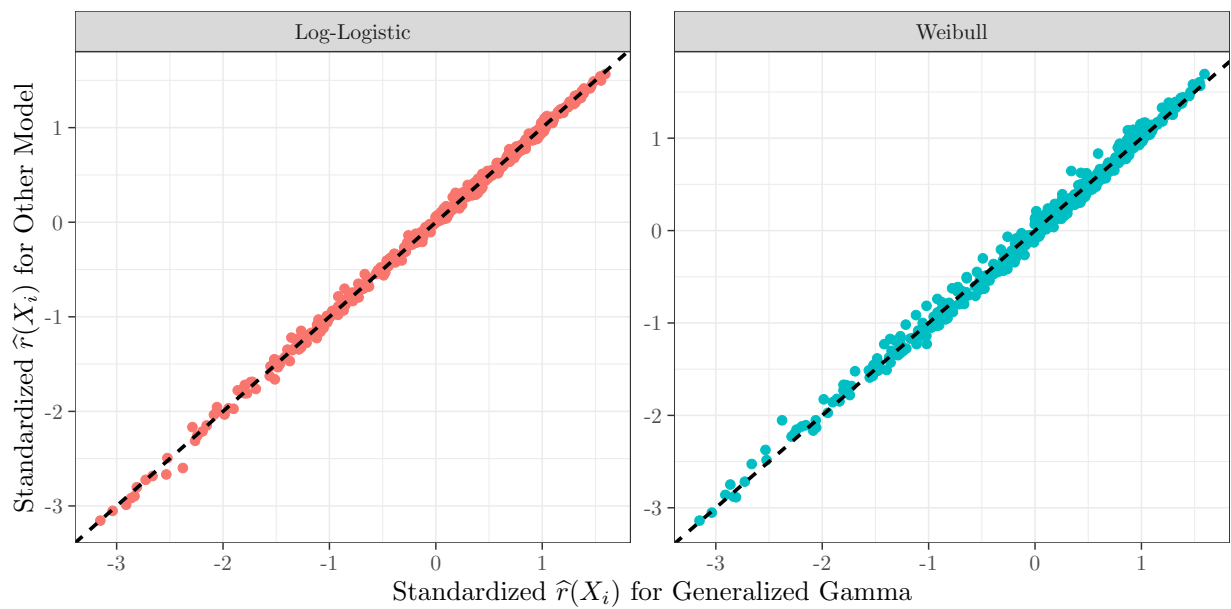

Figure S.5: Comparison of the estimates of $r(X_i)$ for the generalized gamma model to the log-logistic (left) and Weibull (right) models.

## S.2 Justification of our RJMCMC Algorithm

While our algorithm can be described strictly in terms of the RJMCMC framework introduced by Green (1995), we prefer to describe it in a way which keeps the dimension fixed across the transitions; while somewhat unorthodox, we believe that this more easily motivates the particular choices underlying the RJMCMC scheme and makes it more intuitive that the updates are correct.

We associate to $\mathcal{T}$ an infinite branching process prior consisting of nodes $n$ with pairs $(Z_n, \mu_n)$. We imagine generating the tree topology in the following steps.

1. Set $d = 0$ and initialize the tree with a single node $n$ of depth 0.

2. For $d = 0, 1, 2, \ldots$ and each $n$ of depth $d$, sample $\mu_n \sim \pi_\mu$ and a splitting rule $[x_{j_n} \leq C_n]$ and then create two child nodes of depth $d+1$. Then sample $Z_n \sim \text{Bernoulli}(w_n)$ where $w_n = \gamma(1+d)^{-\beta}$ if $Z_b = 1$ for all $b$ on the path from $n$ to the root, and $w_n = 0$ otherwise.

3. For $d = 0, 1, 2, \ldots$ and all $n$ of depth $d$, if $Z_n = 0$ then make $n$ a leaf and remove all nodes which are descendants of $n$; this process terminates if all nodes at depth $d$ are leaves.

A schematic representation of this generative scheme is given in Figure S.6. In the scheme described above, the variable $Z_n$ encodes whether node $n$ is a branch ($Z_n = 1$) or not ($Z_n = 0$). The leaf nodes are precisely those nodes $\ell$ such that $Z_\ell = 0$ but whose parent has $Z_n = 1$. Note also that all infinite latent values of $Z_n$ (but *not* all latent values of $\mu_n$) can be determined from the final decision tree. Rather than having predictions $\mu_\ell$ for $\ell \in \mathcal{L}(\mathcal{T})$ we instead have predictions $\mu_n$ for $n \in \mathcal{N}(\mathcal{T})$. We have $\mu_n \overset{\text{iid}}{\sim} \pi_\mu$ but take $g(x; \mathcal{T}_t, \mathcal{M}_t) = \sum_{\ell \in \mathcal{L}(\mathcal{T}_t)} \mu_\ell I(x \rightsquigarrow \ell)$ so that $\{\mu_b : b \in \mathcal{B}(\mathcal{T}_t)\}$ does not contribute to the sum.

The value of this representation is that it allows us to construct Metropolis-Hastings proposals which do not involve changing the number of continuous parameters; instead, we keep track of the infinite set of both the tree topology $\mathcal{T}_\star$ and an infinite collection of mean

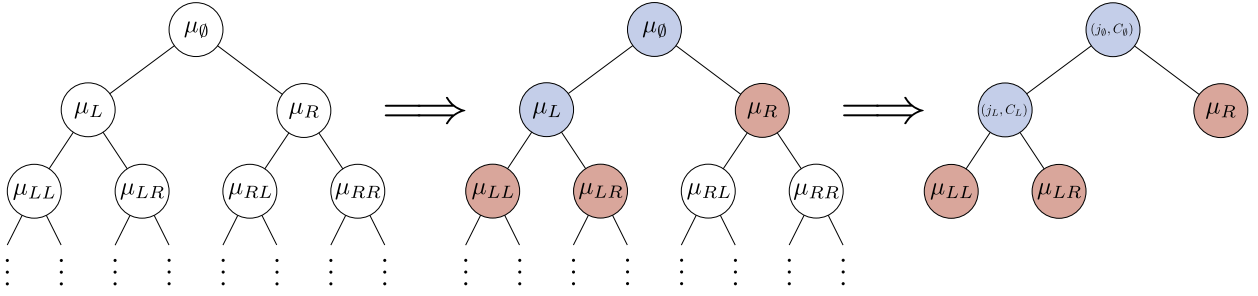

Figure S.6: Sampling $(\mathcal{T}, \mathcal{M})$ from the CART prior using the parameter expanded representation. Blue nodes represent $Z_n = 1$, red nodes represent leaf nodes with $Z_n = 0$, and white nodes represent non-leaf-nodes with $Z_n = 0$.

parameters $\mathcal{M}_\star$. While these structures are infinite, most of their components do not affect the distribution of the observed data, and so have full-conditional equal to their priors; hence, the irrelevant components can be implicitly sampled from their full conditional before each Metropolis-Hastings step, and can be determined *retrospectively* (Papaspiliopoulos and Roberts, 2008) by sampling from their priors whenever they are required in a Metropolis-Hastings step (that is, when we need $\mu_\ell$ we sample it from $\pi_\mu$, and if we need a splitting rule $[x_j \leq C]$ we sample it from its prior).

With this expanded parameter space, we can now describe the reversible jump moves we make to generate proposals $(\mathcal{T}', \mathcal{M}')$ to update $(\mathcal{T}, \mathcal{M})$. Conveniently, while these steps involve introducing auxiliary variables, these auxiliaries end up canceling in the Metropolis-Hastings acceptance ratio, leading to the simplified proposals in Section 3.1.

BIRTH Randomly choose a leaf node $\ell$ of $\mathcal{T}$ and retrospectively sample $\mu_{\ell L} \sim \pi_\mu$, $\mu_{\ell R} \sim \pi_\mu$, and the splitting rule $[x_{j_\ell} \leq C_\ell]$. Set $Z_\ell = 1$, $Z_{\ell L} = 0$, and $Z_{\ell R} = 0$ (converting $\ell$ from a leaf to a branch with two leaf children). Then sample $\mu'_\ell \sim \pi_\mu$ and $(\mu'_{\ell L}, \mu'_{\ell R}) \sim G_{\text{BIRTH}}(\mu'_{\ell L}, \mu'_{\ell R})$.

DEATH Randomly choose a branch node $b \in \text{NOG}(\mathcal{T}_t)$ and retrospectively sample $\mu_b \sim \pi_\mu$. Set $Z_b = 0$, $Z_{bL} = 0$, and $Z_{bR} = 0$ (converting $b$ from a branch to a leaf). Then sample $\mu'_{bL}, \mu'_{bR} \sim \pi_\mu$ and $\mu'_b \sim G_{\text{DEATH}}(\mu')$.

CHANGE Randomly choose a branch node $b \in \text{NOG}(\mathcal{T}_t)$ and sample a new splitting rule $[x_{j'_b} \leq C'_\ell]$ from the prior. Then sample new leaf node predictions $(\mu'_L, \mu'_R) \sim G_{\text{CHANGE}}(\mu'_L, \mu'_R)$.

In the sequel, we derive $R_{\texttt{BIRTH}}$ within this framework.

## S.2.1   Derivation of $R_{\texttt{BIRTH}}$

We give a heuristic derivation of $R_{\texttt{BIRTH}}$ in this section; strictly speaking, this argument ignores the fact that there is no infinite-dimensional analog of Lebesgue measure, however the argument produces the correct answer (which can be verified by standard RJMCMC arguments). The general form for $R_{\texttt{BIRTH}}$ is

$$\frac{\pi(\mathcal{T}'_\star, \mathcal{M}'_\star)}{\pi(\mathcal{T}_\star, \mathcal{M}_\star)} \times \frac{\mathscr{L}(\mathcal{T}'_\star, \mathcal{M}'_\star)}{\mathscr{L}(\mathcal{T}_\star, \mathcal{M}_\star)} \times \frac{Q(\mathcal{T}_\star, \mathcal{M}_\star \mid \mathcal{T}'_\star, \mathcal{M}'_\star)}{Q(\mathcal{T}'_\star, \mathcal{M}'_\star \mid \mathcal{T}_\star, \mathcal{M}_\star)}$$

where $Q(\mathcal{T}_\star, \mathcal{M}_\star \mid \mathcal{T}'_\star, \mathcal{M}'_\star)$ denotes the density of the transition from $(\mathcal{T}'_\star, \mathcal{M}'_\star)$ to $(\mathcal{T}_\star, \mathcal{M}_\star)$. Here, $\mathcal{M}_\star$ denotes the infinite collection of parameters $\{\mu_n : n \in \mathcal{E}\}$ where $\mathcal{E}$ is the collection of all finite strings of $L$'s and $R$'s; similarly, $\mathcal{T}_\star$ is the infinite tree with an infinite collection of splitting rules.

We compute each of the ratios above in turn. First, from our definition of the prior we have

$$\pi(\mathcal{T}_\star, \mathcal{M}_\star) = \prod_{b \in \mathcal{B}(\mathcal{T}_\star)} \rho_{\text{depth}(b)} \prod_{\ell \in \mathcal{L}(\mathcal{T}_\star)} [1 - \rho_{\text{depth}(\ell)}] \prod_{n \in \mathcal{E}} \pi_\mu(\mu_n) \frac{s_{j_n}}{U_{j_n} - L_{j_n}}.$$

Taking the ratio of $\pi(\mathcal{T}_\star, \mathcal{M}_\star)$ and $\pi(\mathcal{T}'_\star, \mathcal{M}'_\star)$ and canceling like terms gives

$$\frac{\pi(\mathcal{T}'_\star, \mathcal{M}'_\star)}{\pi(\mathcal{T}_\star, \mathcal{M}_\star)} = \frac{\rho_d (1 - \rho_{d+1})^2 \, \pi_\mu(\mu'_\ell) \, \pi_\mu(\mu'_{\ell L}) \, \pi_\mu(\mu'_{\ell R})}{(1 - \rho_d) \, \pi_\mu(\mu_\ell) \, \pi_\mu(\mu_{\ell L}) \, \pi_\mu(\mu_{\ell R})}.$$

For the second term, we can write

$$\mathscr{L}(\mathcal{T}_\star, \mathcal{M}_\star) = \prod_{\ell \in \mathcal{L}(\mathcal{T}_\star)} \mathscr{F}(\ell \mid \mathcal{T}_\star, \mu_\ell) / \pi_\mu(\mu_\ell).$$

Taking the ratio of $\mathscr{L}(\mathcal{T}'_\star, \mathcal{M}'_\star)$ and $\mathscr{L}(\mathcal{T}_\star, \mathcal{M}_\star)$ and canceling like terms again gives

$$\frac{\mathscr{L}(\mathcal{T}'_\star, \mathcal{M}'_\star)}{\mathscr{L}(\mathcal{T}_\star, \mathcal{M}_\star)} = \frac{\mathscr{F}(\ell L \mid \mathcal{T}'_\star, \mu'_{\ell L}) \, \mathscr{F}(\ell R \mid \mathcal{T}'_\star, \mu_{\ell R}) \, \pi_\mu(\mu_\ell)}{\mathscr{F}(\ell \mid \mathcal{T}_\star, \mu_\ell) \, \pi_\mu(\mu'_{\ell L}) \, \pi_\mu(\mu'_{\ell R})}.$$

Finally, we calculate the transition ratio. First, the transition from $(\mathcal{T}'_\star, \mathcal{M}'_\star)$ to $(\mathcal{T}_\star, \mathcal{M}_\star)$ requires (i) electing to perform a DEATH move, (ii) selecting $\ell$ from $\text{NOG}(\mathcal{T}'_\star)$, and (iii) proposing $(\mu_\ell, \mu_{\ell L}, \mu_{\ell R})$ from $G_{\texttt{DEATH}}(\mu_\ell) \, \pi_\mu(\mu_{\ell L}) \, \pi_\mu(\mu_{\ell R})$. Hence,

$$Q(\mathcal{T}_\star, \mathcal{M}_\star \mid \mathcal{T}'_\star, \mathcal{M}'_\star) = p_{\texttt{DEATH}}(\mathcal{T}'_\star) \, |\text{NOG}(\mathcal{T}'_\star)|^{-1} \, G_{\texttt{DEATH}}(\mu_\ell) \, \pi_\mu(\mu_{\ell L}) \, \pi_\mu(\mu_{\ell R}).$$

Conversely, to transition from $(\mathcal{T}_\star, \mathcal{M}_\star)$ to $(\mathcal{T}'_\star, \mathcal{M}'_\star)$ we must (i) elect to perform a BIRTH move, (ii) select $\ell$ from $\mathcal{L}(\mathcal{T}_\star)$, and (iii) propose $(\mu'_\ell, \mu'_{\ell L}, \mu'_{\ell R})$ from $G_{\texttt{BIRTH}}(\mu_{\ell L}, \mu_{\ell R}) \, \pi_\mu(\mu_\ell)$. Hence,

$$Q(\mathcal{T}'_\star, \mathcal{M}'_\star \mid \mathcal{T}'_\star, \mathcal{M}'_\star) = p_{\texttt{BIRTH}}(\mathcal{T}_\star) \, |\mathcal{L}(\mathcal{T}'_\star)|^{-1} \, G_{\texttt{BIRTH}}(\mu_{\ell L}, \mu_{\ell R}) \, \pi_\mu(\mu_\ell).$$

The transition ratio is therefore

$$\frac{Q(\mathcal{T}_\star, \mathcal{M}_\star \mid \mathcal{T}'_\star, \mathcal{M}'_\star)}{Q(\mathcal{T}'_\star, \mathcal{M}'_\star \mid \mathcal{T}'_\star, \mathcal{M}'_\star)} = \frac{p_{\texttt{DEATH}}(\mathcal{T}'_\star) \, |\text{NOG}(\mathcal{T}'_\star)|^{-1}}{p_{\texttt{BIRTH}}(\mathcal{T}_\star) \, |\mathcal{L}(\mathcal{T}'_\star)|^{-1}} \times \frac{G_{\texttt{DEATH}}(\mu_\ell) \, \pi_\mu(\mu_{\ell L}) \, \pi_\mu(\mu_{\ell R})}{G_{\texttt{BIRTH}}(\mu'_{\ell L}, \mu'_{\ell R}) \, \pi_\mu(\mu'_\ell)}.$$

Multiplying the three terms together and canceling like terms gives the formula for $R_{\texttt{BIRTH}}$ from Proposition 1.

## S.3 Derivation of $U_\eta(y \mid \lambda)$ and $\mathcal{I}_\eta(\lambda)$ for Structured Variance Regression

Consider the model $Y_i \sim \text{Normal}\{m, \phi V(m)\}$ where $m = g(\lambda)$ and $\eta = (\phi, V(\cdot))$. The log-likelihood for a single observation is

$$\log f_\eta(y \mid \lambda) = -\frac{1}{2}\log(2\pi\phi) - \frac{1}{2}\log V(m) - \frac{(y-m)^2}{2\phi V(m)}.$$

Using the chain rule $\frac{\partial}{\partial \lambda} = g'(\lambda)\frac{\partial}{\partial m}$, the derivative with respect to $\lambda$ is

$$\left(-\frac{V'(m)}{2V(m)} + \frac{V'(m)(y-m)^2}{2\phi V(m)^2} + \frac{y-m}{\phi V(m)}\right) g'(\lambda).$$

Using the product rule, the second derivative is the sum of two components, the first being

$$\left(-\frac{V'(m)}{2V(m)} + \frac{V'(m)(y-m)^2}{2\phi V(m)^2} + \frac{y-m}{\phi V(m)}\right) g''(\lambda).$$

Because the score has mean 0, this term does not appear in $\mathcal{I}_\eta(\lambda)$. The second component is

$$\left(\frac{\partial}{\partial m} - \frac{V'(m)}{2V(m)} + \frac{V'(m)(y-m)^2}{2\phi V(m)^2} + \frac{y-m}{\phi V(m)}\right) g'(\lambda)^2$$

which evaluates to

$$\left(-\frac{V(m)V''(m) - V'(m)^2}{2V(m)^2} \right.$$
$$+ \frac{V''(m)V(m)^{-2}(y-m)^2 - 2V'(m)^2(y-m)^2 V(m)^{-3} - 2V'(m)(y-m)V(m)^{-2}}{2\phi}$$
$$\left. - \frac{V(m)^{-1} + (y-m)V(m)^{-2}V'(m)}{\phi}\right) g'(\lambda)^2.$$

9

Integrating with respect to $f_\eta(y \mid \lambda)$ gives

$$\left(-\frac{V(m)V''(m) - V'(m)^2}{2V(m)^2} + \frac{V''(m)V(m)^{-1} - 2V'(m)^2 V(m)^{-2}}{2} - \frac{V(m)^{-1}}{\phi}\right) g'(\lambda)^2$$
$$= -\left(\frac{V'(m)^2}{2V(m)^2} + \frac{1}{\phi V(m)}\right) g'(\lambda)^2.$$

Summarizing, the Fisher information is given by $\mathcal{I}_\eta(\lambda) = \left(\frac{V'(m)^2}{2V(m)^2} + \frac{1}{\phi V(m)}\right) g'(\lambda)^2$.

Finally, let $\tau = \phi^{-1}$. Then the likelihood is given by

$$\prod_i \sqrt{\frac{\tau}{2\pi}} \exp\left\{-\frac{\tau}{2}\left(\frac{Y_i - m_i}{\sqrt{V(m_i)}}\right)^2\right\} \propto \tau^{-N/2} \exp\left\{-\tau \sum_i \frac{(Y_i - m_i)^2}{2V(m_2)}\right\}$$
$$\propto \text{Gam}\{\tau \mid N/2, 1/2 \sum_i (Y_i - m_i)^2 / V(m_i)\}.$$



\end{document}